%% file: main.tex
\documentclass[twocolumn]{autart}
\usepackage[T1]{fontenc}
\usepackage[english]{babel}
\usepackage[pdftex]{graphicx}   
\usepackage{epsfig}             
\usepackage{amsmath}            
\usepackage{amssymb}            
\usepackage{mathtools}
\usepackage[numbers]{natbib}
\usepackage[bookmarks=true]{hyperref}
\usepackage[x11names]{xcolor}
\usepackage{siunitx}
\usepackage{lipsum}
\usepackage{empheq}
\usepackage[customcolors,markings]{hf-tikz}
\usepackage{nicematrix}
\usepackage[caption=false]{subfig}
\usepackage[commandnameprefix=ifneeded, final]{changes}

\def\preprint{}

\usetikzlibrary{calc,backgrounds}

\usepackage{bm}

\input{preamble/preamble}

\input{preamble/notation_defs}

\input{styles/math}
\input{styles/robotics}
\input{styles/references}

\usepackage{acro}
\input{acro/acro.sty}

\usepackage{tabularray}
\UseTblrLibrary{booktabs}
\usepackage{array}
\newcommand\Tstrut{\rule{0pt}{3.5ex}}
\newcommand\Bstrut{\rule[-1.5ex]{0pt}{0pt}}

\DeclareSIUnit{\sqrts}{\ensuremath{\sqrt{\text{\second}}}}

\begin{document}
\renewcommand{\baselinestretch}{0.95}

\begin{frontmatter}

\title{Equivariant Symmetries for Inertial Navigation Systems}

\author[CNS]{Alessandro Fornasier\corauthref{corauth}}\ead{alessandro.fornasier@aau.at},
\author[ANU]{Yixiao Ge}\ead{yixiao.ge@anu.edu.au},
\author[ANU]{Pieter van Goor}\ead{pieter.vangoor@anu.edu.au},
\author[ANU]{Robert Mahony}\ead{robert.mahony@anu.edu.au},
\author[CNS]{Stephan Weiss}\ead{stephan.weiss@aau.at}

\address[CNS]{Control of Networked Systems Group, University of Klagenfurt, Austria}
\address[ANU]{System Theory and Robotics Lab, Australian Centre for Robotic Vision, Australian National University, Australia}

\corauth[corauth]{Corresponding author}


\input{sections/abstract}

\end{frontmatter}

\input{sections/introduction}
\input{sections/notation}
\input{sections/outline}
\input{sections/system}
\input{sections/symmetries}
\input{sections/linearisation}
\input{sections/experiments}
\input{sections/conclusion}
\ifdefined\preprint
\appendix
\input{sections/appendix}
\fi

\bibliographystyle{plain}


\end{document}

%% file: preamble/preamble.tex
\usepackage[T1]{fontenc}

\usepackage{MnSymbol}
\usepackage{mathdots}

%% file: preamble/notation_defs.tex
\providecommand{\R}{\mathbb{R}}

\providecommand{\SO}{\mathbf{SO}}

\providecommand{\SE}{\mathbf{SE}}

\providecommand{\HG}{\mathbf{HG}}

\providecommand{\grpG}{\mathbf{G}}
\providecommand{\grpH}{\mathbf{H}}

\providecommand{\grpA}{\mathbf{G_{DP}}}
\providecommand{\grpB}{\mathbf{G_{SD}}}
\providecommand{\grpC}{\mathbf{G_{TF}}}
\providecommand{\grpD}{\mathbf{G_{TG}}}
\providecommand{\grpE}{\mathbf{G_{O}}}
\providecommand{\grpF}{\mathbf{G_{ES}}}

\providecommand{\gothso}{\mathfrak{so}}
\providecommand{\gothse}{\mathfrak{se}}

\providecommand{\gothg}{\mathfrak{g}}

\providecommand{\gothX}{\mathfrak{X}} 

\providecommand{\gothhg}{\mathfrak{hg}}

\providecommand{\gothGrpA}{\mathfrak{g}_{\mathbf{DP}}}

\providecommand{\gothGrpD}{\mathfrak{g}_{\mathbf{TG}}}
\providecommand{\gothGrpE}{\mathfrak{g}_{\mathbf{O}}}
\providecommand{\gothGrpF}{\mathfrak{g}_{\mathbf{ES}}}
\providecommand{\so}{\mathfrak{so}}
\providecommand{\se}{\mathfrak{se}}

\providecommand{\sdpgrpG}{\mathbf{G}_{\mathfrak{g}}^{\ltimes}}

\providecommand{\calG}{\mathcal{G}}

\providecommand{\calM}{\mathcal{M}}
\providecommand{\calN}{\mathcal{N}}
\providecommand{\calO}{\mathcal{O}}

\providecommand{\calHG}{\mathcal{HG}}

\providecommand{\torSO}{\mathcal{SO}}

\providecommand{\torSE}{\mathcal{SE}}

\providecommand{\vecL}{\mathbb{L}}

\DeclareMathOperator{\Ad}{Ad}
\DeclareMathOperator{\ad}{ad}

\providecommand{\tT}{\mathrm{T}} 
\providecommand{\td}{\mathrm{d}}
\providecommand{\tD}{\mathrm{D}}

\providecommand{\ddt}{\frac{\td}{\td t}}

\providecommand{\Fr}[2]{\left. \mathrm{D}_{#1} \right|_{#2}}

\usepackage{accents}
\makeatletter
\providecommand{\scirc}{%
    \hbox{\fontfamily{\rmdefault}\fontsize{0.4\dimexpr(\f@size pt)}{0}\selectfont{\raisebox{-0.52ex}[0ex][-0.52ex]{$\circ$}}}}

\makeatother

\mathchardef\mhyphen="2D



%% file: styles/math.tex
\newcommand{\st}{\;|\;}

\newcommand{\AtoB}[2]{\;:\;#1\;\rightarrow\;#2}

\newcommand{\chrulefill}{%
  \leaders\hrule height \dimexpr\fontdimen22\scriptscriptfont2+0.2pt\relax
                 depth -\dimexpr\fontdimen22\scriptscriptfont2-0.2pt\relax
          \hfill}
\newcommand{\xoverline}[2]{%
  \mathord{
    \vbox{\offinterlineskip
      \halign{##\cr
        \chrulefill$\,\scriptscriptstyle#1\,$\chrulefill\cr
        \noalign{\kern.2ex}
        $#2$\cr
      }%
    }%
  }%
}

\newcommand{\set}[2]{\left\{ #1 \;\middle|\; #2 \right\}}

\providecommand{\tT}{\mathrm{T}} 
\providecommand{\td}{\mathrm{d}}
\providecommand{\tD}{\mathrm{D}}

\providecommand{\ddt}{\frac{\td}{\td t}}

\providecommand{\Fr}[2]{\mathrm{D}_{#1}\big|_{#2}}

\newcommand{\eye}{\mathbf{I}}

\newcommand{\Adsym}[2]{\mathrm{Ad}_{#1}\left[#2\right]}
\newcommand{\adsym}[2]{\mathrm{ad}_{#1}\left[#2\right]}

\newcommand{\AdMsym}[1]{\mathbf{Ad}^\vee_{#1}}
\newcommand{\adMsym}[1]{\mathbf{ad}^\vee_{#1}}

\providecommand{\xizero}{\mathring{\xi}}
\providecommand{\yzero}{\mathring{y}}

%% file: styles/robotics.tex
\newcommand{\Vector}[3]{\prescript{#1}{}{\bm{#2}}_{#3}}
\newcommand{\dotVector}[3]{\prescript{#1}{}{\dot{\bm{#2}}}_{#3}}
\newcommand{\hatVector}[3]{\prescript{#1}{}{\hat{\bm{#2}}}_{#3}}

\newcommand{\Pose}[2]{\prescript{#1}{}{\mathbf{T}}_{#2}}

\newcommand{\Rot}[2]{\prescript{#1}{}{\mathbf{R}}_{#2}}

\newcommand{\hatPose}[2]{\prescript{#1}{}{\hat{\mathbf{T}}_{#2}}}

\newcommand{\hatRot}[2]{\prescript{#1}{}{\hat{\mathbf{R}}_{#2}}}

\newcommand{\ubarVector}[3]{\prescript{#1}{}{\overline{\bm{#2}}}_{#3}}
\newcommand{\lbarVector}[3]{\prescript{#1}{}{\underline{\bm{#2}}}_{#3}}

\newcommand{\frameofref}[1]{\{$#1$\}}

%% file: styles/references.tex
\newcommand{\figref}[1]{Fig.~\ref{fig:#1}}
\newcommand{\secref}[1]{Sec.~\ref{sec:#1}}
\newcommand{\equref}[1]{Equ.~(\ref{eq:#1})}

\newcommand{\tabref}[1]{Tab.~\ref{tab:#1}}
\newcommand{\cit}[1]{~\cite{#1}}
\providecommand{\etal}{\textit{et al.}}

%% file: sections/abstract.tex
\begin{abstract}
This paper investigates the problem of \ac{ins} filter design through the lens of symmetry. 
The \ac{ekf} and its variants have been the staple of \ac{ins} filtering for 50 years. However, recent advances in \aclp{ins} have exploited matrix Lie group structure to design stochastic filters and state observers that have been shown to display superior performance compared to classical solutions.
In this work, we explore various symmetries of \acl{ins}, including two novel symmetries that have not been considered in the prior literature, and provide a discussion of the relative strengths and weaknesses of these symmetries in the context of filter design.
We show that all the modern variants of the \ac{ekf} for inertial navigation can be interpreted as the recently proposed \ac{eqf} design methodology applied to different choices of symmetry group for the \ac{ins} problem. 
As a direct application of the symmetries presented, we address the filter design problem for a vehicle equipped with an \ac{imu} and a \ac{gnss} receiver, providing a comparative analysis of different modern filter solutions.
We believe the collection of symmetries that we present here capture all the sensible choices of symmetry for this problem, and that the analysis provided is indicative of the relative real-world performance potential of the different algorithms for trajectories ensuring full state observability.
\end{abstract}

%% file: sections/introduction.tex
\section{Introduction}
The theory of invariant filtering for group affine systems~\cite{Barrau2022TheProblems, barrau:tel-01247723} and the theory of equivariant filters~\cite{Mahony2020EquivariantDesign, Ng2020EquivariantGroups, VanGoor2020EquivariantSpaces} that generalizes to systems on homogeneous spaces have provided general design frameworks, as well as strong theoretical performance guarantees, for filter designs that exploit symmetry. 
This has motivated the widespread use of invariant filters in the robotics community, and its adoption for inertial navigation problems~\cite{doi:10.1177/0278364919894385, Pavlasek2021InvariantEstimation, Li2022Closed-FormVINS}. 
The application of these principles to inertial navigation systems (\ac{ins}) has seen the most significant performance gains from algorithm design in this field for the last 40 years.
There are now several competing modern \ac{ins} filters based on geometric insights available in the literature~\cite{7523335, Barrau2022TheProblems, Fornasier2022EquivariantBiases} and the question of how to analyse and evaluate the similarities and differences is now of interest. 
A recent paper by Barrau \etal~states \emph{``The big question when it comes to invariant observers/filters is how do we find a group structure for the state space [...]''}~\cite{Barrau2022TheProblems}. 
The goal of the present paper is to convince the reader that the choice of symmetry structure is in fact the \emph{only} difference between different versions of modern geometric \ac{ins} filters. 

In this paper we present six different symmetry groups that act on the state-space of the \ac{ins} problem. 
We use the recent equivariant filter design methodology to generate \ac{ins} filter algorithms for each of these symmetries. 
We show that the classical \ac{mekf}  \cite{Lefferts1982KalmanEstimation}, the Imperfect-\ac{iekf} \cite{7523335}, the \ac{tfgiekf} \cite{Barrau2022TheProblems}, and the authors own recent work proposing an equivariant filter for the tangent group structure (TG-EqF) \cite{Fornasier2022EquivariantBiases} are all associated with equivariant filter design \cite{vanGoor2022EquivariantEqF} applied to different symmetry actions on the same state-space. 
This leads us to consider the properties of the symmetry groups and suggests two additional symmetries leading to filters, that we term the Direct Position Equivariant Filter (DP-EqF) and Semi-Direct Bias Equivariant Filter (SD-EqF), that are novel and do not correspond to prior algorithms in the literature.  

In the context of \acs{gnss}-based navigation, we derive \ac{eqf} algorithms for all of the different symmetries, demonstrating that this approach provides a unifying analysis framework for modern \ac{ins} filters. 
In doing this we also make a minor contribution in demonstrating how fixed-frame measurements can be reformulated as body-frame relative measurements.  
This allows us to exploit output equivariance~\cite{vanGoor2022EquivariantEqF} for all the filter geometries, ensuring at least third-order linearization error in the output equations. 

We undertake a simple comparative study in concert with a linearization analysis of the error equations. 
We consider the ``fully observable'' case where all states are estimated and are observable for the given system trajectories. 
We recognise that for specific cases where trajectories lead to unobservable states (e.g.,~straight-line flight, hovering, etc.) some of the following observations may not hold.
For such trajectories we make the following observations:
\begin{itemize}
\item 
The classical MEKF demonstrates noticeable performance limitations compared to the modern filters. 
In particular, it demonstrates worse transient response and reports significant overconfidence during the transient phase. 
\item 
The performance differences in modern filters are primarily visible during the transient phase of error response. 
The asymptotic behaviour of all filters is similar. 
\item Notwithstanding the above, the asymptotic performance of the TG-EqF appears superior to all other filters demonstrating the best consistency and the lowest error. 
%
\item The TG-\ac{eqf} filter is the only filter with exact linearization of the navigation states; the only nonlinearities occur in the bias states. The authors believe that this property underlies its performance advantage. 
\item The bias state transient response of the filters with semi-direct bias symmetry (TG-\ac{eqf}, DP-\ac{eqf} and SD-\ac{eqf}) appears superior to that of filters without this geometric structure (\ac{mekf}, \ac{iekf} and \ac{tfgiekf}). 
\end{itemize}

The study concludes that any of the IEKF, TG-EqF, DP-EqF, and SD-EqF filters are candidates for a high-performance \ac{ins} filter design. 
The lower filter error and energy properties of the TG-EqF recommend it as the  leading choice for fully observable trajectories.

%% file: sections/notation.tex
\section{Notation and Preliminaries}\label{sec:prelim}

In this paper bold lowercase letters are used to indicate vector quantities.
Bold capital letters are used to indicate matrices.
Regular letters are used to indicate elements of a symmetry group. 

Frames of reference are denoted as \frameofref{A} and \frameofref{B}.
Vectors describing physical quantities expressed in a frame of reference \frameofref{A} are denoted by $\Vector{A}{x}{}$.
Rotation matrices encoding the orientation of a frame of reference \frameofref{B} with respect to a reference \frameofref{A} are denoted by $\Rot{A}{B}$; in particular, $\Rot{A}{B}$ expresses a vector $\Vector{B}{x}{}$ defined in the \frameofref{B} frame of reference into a vector ${\Vector{A}{x}{} = \Rot{A}{B}\Vector{B}{x}{}}$ expressed in the \frameofref{A} frame of reference. Finally, ${\eye_n \in \R^{n \times n}}$ is the $n \times n$ identity matrix, and ${\mathbf{0}_{n \times m} \in \R^{n \times m}}$ is the $n \times m$ zero matrix.

For all $\Vector{}{x}{} \in \R^n$ define the maps:
\begin{align*}
    &\ubarVector{}{\left(\cdot\right)}{} \AtoB{\R^n}{\R^{n+3}}, \qquad \Vector{}{x}{} \mapsto \ubarVector{}{x}{} = \left(\mathbf{0}_{3 \times 1}, \Vector{}{x}{}\right) ,\\
    &\lbarVector{}{\left(\cdot\right)}{} \AtoB{\R^n}{\R^{n+3}}, \qquad \Vector{}{x}{} \mapsto \lbarVector{}{x}{} = \left(\Vector{}{x}{}, \mathbf{0}_{3 \times 1}\right) . 
\end{align*}
%

The following Lie groups are used throughout the paper.
\begin{align*}
    \SO(3) &= \set{\mathbf{A} \in \R^{3\times3}}{
    \mathbf{A} \mathbf{A}^\top = \mathbf{I}_3, \; \det(\mathbf{A} ) = 1}, \\
    \SE(3) &= \set{
    \begin{bmatrix}
        \mathbf{A} & \Vector{}{b}{} \\ \mathbf{0}_{1\times 3} & 1
    \end{bmatrix}
    \in \R^{4\times4}}{
    \mathbf{A} \in \SO(3), \; \Vector{}{b}{} \in \R^3}, \\
    \SE_2(3) &= \set{
    \begin{bmatrix}
        \mathbf{A} & \Vector{}{a}{} & \Vector{}{b}{} \\ 
        \mathbf{0}_{1\times 3} & 1 & 0\\
        \mathbf{0}_{1\times 3} & 0 & 1
    \end{bmatrix}
    \in \R^{5\times5}}{
    \mathbf{A} \in \SO(3), \; \Vector{}{a}{}, \Vector{}{b}{} \in \R^{3}}.
\end{align*}
Their Lie algebras are the tangent spaces at the identities of each group.

For all $X = \left(\mathbf{A}, \Vector{}{a}{}, \Vector{}{b}{}\right) \in \SE_2(3) \st \mathbf{A} \in \SO(3), \Vector{}{a}{}, \Vector{}{b}{} \in \R^3$, define the map:
\begin{equation*}
    \Omega\left(\cdot\right) \AtoB{\SE_2(3)}{\se_2(3)}, \quad \Omega\left(X\right) = \left(\mathbf{0}_{3 \times 1}, \mathbf{0}_{3 \times 1}, \Vector{}{a}{}\right)^\wedge \in \se_2(3).
\end{equation*}
For all $\Vector{}{p}{}, \Vector{}{q}{}, \Vector{}{r}{} \in \R^3 \st (\Vector{}{p}{}, \Vector{}{q}{}, \Vector{}{r}{}) \in \R^9$, define the map:
\begin{equation*}
    \Pi\left(\cdot\right) \AtoB{\se_2(3)}{\se(3)}, \quad \Pi\left(\left(\Vector{}{p}{}, \Vector{}{q}{}, \Vector{}{r}{}\right)^{\wedge}\right) = \left(\Vector{}{p}{}, \Vector{}{q}{}\right)^{\wedge} \in \se(3).
\end{equation*}


In what follows, we recall the concept of symmetry, equivariance, and \acl{eqf} design. For an introduction to Lie groups and homogeneous spaces in the context of \acl{eqf}, we refer the reader to the authors' prior work~\cite{Mahony2020EquivariantDesign, Mahony2022ObserverEquivariance, vanGoor2022EquivariantEqF, Fornasier2022EquivariantBiases}. 
The
\ifdefined\preprint
appendix~\ref{sec:append}
\else
extended preprint version of this paper~\cite{Fornasier2023EquivariantSystems} 
\fi
also provides additional explanation. 

\subsection{Symmetry Equivariance and Lifted System}
\label{sec:symmetry_definitions}

A symmetry of a kinematic system can be seen as a set of transformations that either leave unchanged or change in a structured manner the equations that govern the motion of the system.
This is encoded by a transitive group action $\phi$ of a Lie group $\grpG$ (also called a \emph{symmetry group}) on the state space $\calM$ of a system.
Formally, a (right) Lie group action $\phi : \grpG \times \calM \to \calM$ is a smooth map satisfying
\begin{align*}
    \phi(XY, \xi) = \phi(Y, \phi(X, \xi)), &&
    \phi(I, \xi) = \xi,
\end{align*}
for all $X,Y \in \grpG$ and $\xi \in \calM$.
An action $\phi$ is said to be \emph{transitive} if for all $\xi_1,\xi_2 \in \calM$, there exists $X \in \grpG$ such that $\phi(X, \xi_1) = \xi_2$.

A symmetry that transforms the equations of motion of a kinematic system in a structured manner encodes \emph{\deleted{equivaraince}\added{equivariance}} of the system.
Formally, consider a system $f : \vecL \to \gothX(\calM)$ as a map from a vector space of inputs $\vecL$ to vector fields on the state space $\calM$; that is, each input $u \in \vecL$ corresponds to a vector field $f_u \in \gothX(\calM)$ on the state space $\calM$.
Then $f$ is said to be \emph{equivariant} if
\begin{equation}\label{eq_equi}
    \td\phi_{X} \circ f_u \circ \phi_{X^{-1}} = f_{\psi_X(u)},
\end{equation}
$\forall X \in \grpG$, $u \in \vecL$, and for a right-handed group action $\psi \AtoB{\grpG \times \vecL}{\vecL}$ of the group $\grpG$ on the input space $\vecL$.
Similarly, a symmetry that changes the output in a structured manner encodes  \emph{equivariance of the output}.
Formally, consider an output map $h:\calM \to \calN$ where $\calN$ is a smooth manifold called the output space.
Then $h$ is said to be \emph{equivariant} if
\begin{equation}\label{eq_out_equi}
    h(\phi_X(\xi)) = \rho_X(h(\xi)) ,
\end{equation}
$\forall X \in \grpG$, $\xi \in \calM$, and for a right-handed action of the group $\grpG$ on the output space $\calN$, that is $\rho \AtoB{\grpG \times \calN}{\calN}$.

Note that for any fixed $X \in \grpG$ the actions $\phi, \psi$ and $\rho$ can be written as diffeomorphisms $\phi_X \AtoB{\calM}{\calM}$, $\psi_X \AtoB{\vecL}{\vecL}$ and $\rho_X \AtoB{\calN}{\calN}$.

If a system possesses a symmetry $(\grpG, \phi)$, the Lie group structure of the symmetry group can be exploited to ``lift'' the system dynamics on the Lie group.
That is, with an arbitrary but fixed choice of origin point $\xizero\in\calM$, defining a geometric structure called ``Lift'' $\Lambda \AtoB{\calM \times \vecL}{\gothg}$ and defining a \emph{lifted system} ${\dot{X} = X\Lambda\left(\phi_X(\xizero), u\right)}$ whose solutions $X(t)$ at time $t$, project to solutions $\xi(t)$~\cite{Mahony2020EquivariantDesign, vanGoor2022EquivariantEqF}.

\subsection{Equivariant Filter Design}

For state estimation problems, lifting the system dynamics onto the Lie group translates the problem to that of estimating an element of the symmetry group ${\hat{X} \in \grpG}$ such that  ${\hat{\xi} = \phi(\hat{X}, \xizero)}$, rather than ${\hat{\xi} \in \calM}$ directly.
This is not just an ``embedding'' of the system on the Lie group, but abstracts the estimation problem to the Lie group.
Exploiting the symmetry of the kinematic systems allows the definition of a \emph{global error} ${e = \phi(\hat{X}^{-1}, \xi) \in \calM}$, termed \emph{equivariant error}~\cite{vanGoor2022EquivariantEqF, VanGoor2023EquivariantAwareness}.
The \ac{eqf} is then the state estimation algorithm that results from applying \ac{ekf} design principles to the (global) error kinematics,  linearized about the fixed origin $\xizero$.

Let $f : \vecL \to \gothX(\calM)$ be a system whose state space is a $m$-dimensional homogeneous space $\calM$, and let $h : \calM \to \calN$ be an output function where the output space is a $n$-dimensional smooth manifold $\calN$.
Assume that $f$ and $h$ are equivariant as in Section~\ref{sec:symmetry_definitions}.
Let $\xi(t) \in \calM$ be a trajectory of the system with measurements,
\begin{align*}
    \dot{\xi} &= f_u(\xi), &
    y(t_k) &= h(\xi(t_k)),
\end{align*}
where $u(t) \in \vecL$ is a measured input signal, and the measurements $y(t_k) \in \calN$ occur at discrete times $t_1 < t_2 < \cdots$.
Choose a fixed origin $\xizero \in \calM$ and let $\Lambda \AtoB{\calM \times \vecL}{\gothg}$ be a lift of the system.
Let $\hat{X} \in \grpG$ denote the observer state and define its dynamics to be
\begin{align*}
    \dot{\hat{X}} &= \hat{X} \Lambda(\phi(\hat{X}, \xizero), u), &
    \hat{X}(t_k^+) &= \exp(\Delta(t_k)) \hat{X}(t_k^+),
\end{align*}
where the $\Delta(t_k) \in \gothg$ are correction terms that depend on the measurements $y(t_k)$.

The \emph{equivariant error} is defined to be $e \coloneqq \phi_{\hat{X}^{-1}}\left(\xi\right)$.
Choose local coordinates of the state space ${\vartheta \AtoB{\mathcal{U}_{\xizero}}{\R^{m}}}$ in a neighbourhood $\mathcal{U}_{\xizero} \subset \calM$ of $\xizero$, and choose local coordinates of the output space ${\delta \AtoB{\mathcal{U}_{\yzero}}{\R^{n}}}$ in a neighbourhood $\mathcal{U}_{\yzero} \subset \calN$ of $\yzero = h(\xizero)$.
Let $\varepsilon = \vartheta(e)$ denote the local coordinates of the error $e$. 
Then the linearized error dynamics and linearized output are~\cite{VanGoor2020EquivariantSpaces}
\begin{align*}
    &\dot{\varepsilon} \approx \mathbf{A}_{t}^{0}\varepsilon 
    ,\\
    &\mathbf{A}_{t}^{0} = \Fr{e}{\xizero}\vartheta\left(e\right)\Fr{E}{I}\phi_{\xizero}\left(E\right)\Fr{e}{\xizero}\Lambda\left(e, \mathring{u}\right)\Fr{\varepsilon}{\mathbf{0}}\vartheta^{-1}\left(\varepsilon\right) ,\\
    &\delta\left(h\left(e\right)\right) = \delta\left(h\left(\rho(\hat{X}^{-1}, y)\right)\right) \approx \mathbf{C}^{0}\bm{\varepsilon} , \\
    &\mathbf{C}^{0} = \Fr{y}{\yzero}\delta\left(y\right)\Fr{e}{\xizero}h\left(e\right)\Fr{\bm{\varepsilon}}{\mathbf{0}}\vartheta^{-1}\left(\bm{\varepsilon}\right) .
\end{align*}
If no compatible action $\psi$ of the symmetry group on the input space is found, the state matrix can be computed alternatively according to
\begin{align*}
    &\mathbf{A}_{t}^{0} = \Fr{e}{\xizero}\vartheta\left(e\right)
    \Fr{\xi}{\hat{\xi}}\phi_{\hat{X}^{-1}}\left(\xi\right)
    \Fr{E}{I}\phi_{\hat{\xi}}\left(E\right)\\
    &\qquad\;\Fr{\xi}{\phi_{\hat{X}}\left(\xizero\right)}\Lambda\left(\xi, u\right)
    \Fr{e}{\xizero}\phi_{\hat{X}}\left(e\right)
    \Fr{\Vector{}{\varepsilon}{}}{\mathbf{0}}\vartheta^{-1}\left(\Vector{}{\varepsilon}{}\right).
\end{align*}
Output equivariance can be exploited to derive a linearized output with third order error~\cite{vanGoor2022EquivariantEqF} as follows
\begin{align*}
    &\delta\left(h\left(e\right)\right) = \delta\left(\rho_{\hat{X}^{-1}}\left(h\left(\xi\right)\right)\right) \approx \mathbf{C}^{\star}\bm{\varepsilon} + \mathbf{O}(\bm{\varepsilon}^3),\\
    &\mathbf{C}^{\star}\bm{\varepsilon} = \frac{1}{2}\Fr{y}{\yzero}\delta\left(y\right)\left(\Fr{E}{I}\rho_E\left(\yzero\right) + \Fr{E}{I}\rho_E\left(\rho_{\hat{X}^{-1}}\left(y\right)\right)\right)\varepsilon^{\wedge}.
\end{align*}

Let $\mathbf{C}$ be either $\mathbf{C}^{0}$ or $\mathbf{C}^{\star}$, then the \ac{eqf} algorithm is
\begin{align*}
\intertext{\textbf{Predict:}}
    &\dot{\hat{X}} = \td \textrm{L}_{\hat{X}}\Lambda(\phi_{\xizero}(\hat{X}), u),\\
    &\dot{\bm{\Sigma}} = \mathbf{A}_{t}^{0}\bm{\Sigma} + \bm{\Sigma}{\mathbf{A}_{t}^{0}}^\top + \mathbf{Q},\\
\intertext{\textbf{Update:}}
    &\Delta = \Fr{E}{I}\phi_{\xizero}\left(E\right)^{\dagger}\td\vartheta^{-1}\bm{\Sigma}{\mathbf{C}}^\top(\mathbf{C}\bm{\Sigma}{\mathbf{C}}^\top + \mathbf{R})^{-1}\delta(\rho(\hat{X}^{-1}, y)) ,\\
    &\bm{\Sigma} = \left(\eye - \bm{\Sigma}{\mathbf{C}}^\top(\mathbf{C}\bm{\Sigma}{\mathbf{C}}^\top + \mathbf{R})^{-1}\mathbf{C}\right)\bm{\Sigma},\\
    &\hat{X} = \exp\left(\Delta\right)\hat{X},\\
\intertext{\textbf{Reset:}}
    &\bm{\Sigma} = \exp\left(\bm{\Gamma}_{\td\phi_{\xizero}\Delta}\right)\bm{\Sigma}\exp\left(\bm{\Gamma}_{\td\phi_{\xizero}\Delta}\right)^\top.
\end{align*}
where the last equation (Reset) accounts for the distortion of the covariance due to the change of coordinate maps on a non-flat manifold, and is often referred to as \emph{reset step} or \emph{curvature correction}~\cite{Mahony2022ObserverEquivariance, Ge2024AGroups}. 

The equations above resemble those of continuous-discrete Kalman-like filters, and indeed, the \ac{eqf} has the same order of computational complexity as any other Kalman-like filter, such as \acs{ekf} and \acs{mekf}.

%% file: sections/outline.tex
\section{Outline of the Paper}
The paper is organized as follows. Section~~\ref{sec:sys} introduces the biased \acl{ins} considered. 
Section~~\ref{sec:sym} introduces and analyzes different symmetries of the biased inertial navigation system under the lens of equivariance. That is, for each symmetry, its equivariance properties, as well as the relation to classical filter design when exploited within the \acl{eqf} framework, are discussed. 
In particular, Tab.~\ref{tab:symmetries_overview} provides an summary of the different symmetries considered, with its rightmost column showing the linearized error dynamics when these symmetries are exploited for filter design. 
Section~~\ref{sec:le_analysis} describes how the linearization error analysis is carried out. 
For readers who do not wish to replicate the straightforward but tedious calculations required to compute the linearisations, the detailed derivations are provided in the appendix 
\ifdefined\preprint
\ref{sec:append}.
\else
of the extended preprint of this manuscript~\cite[Appendix A]{Fornasier2023EquivariantSystems}.
\fi
Section~~\ref{sec:appl} discussed the problem of \ac{uav} position-based localization as a direct application of the symmetries presented in Section~~\ref{sec:sym}. This application serves as a convenient framework to introduce an interesting result in Section~~\ref{sec:reformulation}. That is, how global-referenced measurements are reformulated as residual body-referenced measurements that are compatible with the presented symmetries.
Finally, Section~~\ref{sec:appl} concludes with an experimental validation of the performance of the different \aclp{eqf} built upon the symmetries discussed in Section~~\ref{sec:sym}.

%% file: sections/system.tex
\section{The Biased Inertial Navigation Problem}\label{sec:sys}

Consider a mobile robot equipped with an \ac{imu} providing angular velocity and acceleration measurements, as well as other sensors providing partial direct or indirect state measurements (e.g. a \ac{gnss} receiver providing position measurements or a magnetometer providing direction measurements).
Let \frameofref{G} denote the global inertial frame of reference and \frameofref{I} denote the \ac{imu} frame of reference. In non-rotating, flat earth assumption, the deterministic (noise-free) continuous-time biased inertial navigation system is
\begin{subequations}\label{eq:bins}
    \begin{align}
        &\dot{\Rot{G}{I}} = \Rot{G}{I}\left(\Vector{I}{\bm{\omega}}{} - \Vector{I}{b}{\bm{\omega}}\right)^{\wedge} ,\\
        &\dotVector{G}{v}{I} =  \Rot{G}{I}\left(\Vector{I}{a}{} - \Vector{I}{b}{a}\right) + \Vector{G}{g}{} ,\\
        &\dotVector{G}{p}{I} = \Vector{G}{v}{I} ,\label{eq:bins_p}\\
        &\dotVector{I}{b_{\bm{\omega}}}{} = \Vector{I}{\tau}{\bm{\omega}} ,\\
        &\dotVector{I}{b_{a}}{} = \Vector{I}{\tau}{a} .
    \end{align}
\end{subequations}
Here, $\Rot{G}{I}$ denotes the rigid body orientation, and $\Vector{G}{p}{I}$ and $\Vector{G}{v}{I}$ denote the rigid body position and velocity expressed in the \frameofref{G} frame, respectively.
These variables are termed the \emph{navigation states}.
The gravity vector $\Vector{G}{g}{}$ is expressed in frame \frameofref{G}. 
The gyroscope measurement and accelerometer measurement are written $\Vector{I}{\bm{\omega}}{}$ and $\Vector{I}{\bm{a}}{}$ respectively. 
The two biases $\Vector{I}{b_{\bm{\omega}}}{}$ and $\Vector{I}{b_{\bm{a}}}{}$ are termed the \emph{bias states}. 
The inputs $\bm{\tau_{\omega}}$, $\bm{\tau_{a}}$ are used to model the biases' dynamics, and are zero when the biases are modeled as constant quantities.

The state space is $\calM = \torSO(3) \times \R^3 \times \R^3 \times \R^3 \times \R^3$ where the 4 copies of $\R^3$ model velocity, position, and angular and acceleration bias, respectively, and $\torSO(3)$ is the $\SO(3)$-torsor with rotation matrices representing coordinates of orientation rather than physical rotation of space. 
Note that the state space itself is not a Lie-group in the EqF formulation. 
Rather symmetry is modeled as a group action on $\calM$, allowing us to consider different symmetries acting on the same INS state. 
We write an element of the state space, and an element of the input space respectively
\begin{align}
    &\xi =  \left(\Rot{G}{I}, \Vector{G}{v}{I}, \Vector{G}{p}{I}, \Vector{I}{b_{\bm{\omega}}}{}, \Vector{I}{b_{a}}{} \right) \in \calM ,\\
    & u = \left(\Vector{I}{\bm{\omega}}{}, \Vector{I}{a}{}, \Vector{I}{\tau}{\bm{\omega}}, \Vector{I}{\tau}{a}\right) \in \vecL \subset \R^{12} .
\end{align}

For the sake of clarity of the presentation, in the following sections, we drop subscripts and superscripts from state, input and output variables, and adopt the lean notation defined in Table \ref{tab:conversion}.

\begin{table}[t]
    \renewcommand{\arraystretch}{1.5}
    \setlength\tabcolsep{5pt}
    \centering
    \caption{Descriptive-Lean Notation Conversion Table.}
    \begin{tblr}[
        caption = {Descriptive-Lean Notation Conversion Table.},
    ]{
        rows = {m},
        column{1} = {4.0cm, c},
        column{2} = {1.5cm, c},
        column{3} = {1.5cm, c},
    }
    \toprule
    Description & Descriptive notation & Lean notation \\
    \midrule
    Rigid body orientation & $\Rot{G}{I}$ & $\Rot{}{}$ \\
    Rigid body velocity & $\Vector{G}{v}{I}$ & $\Vector{}{v}{}$ \\
    Rigid body position & $\Vector{G}{p}{I}$ & $\Vector{}{p}{}$ \\
    Angular velocity measurement & $\Vector{I}{\bm{\omega}}{}$ & $\Vector{}{\bm{\omega}}{}$ \\
    Gyroscope bias & $\Vector{I}{b}{\bm{\omega}}$ & $\Vector{}{b}{\bm{\omega}}$ \\
    Acceleration measurement & $\Vector{I}{a}{}$ & $\Vector{}{a}{}$ \\
    Accelerometer bias & $\Vector{I}{b}{\bm{a}}$ & $\Vector{}{b}{\bm{a}}$ \\
    \bottomrule
    \end{tblr}
    \label{tab:conversion}
\end{table}

%% file: sections/symmetries.tex
\begin{table*}
    \renewcommand{\arraystretch}{1.5}
    \setlength\tabcolsep{5pt}
    \centering
    \caption{Qualitative overview of the differences in the presented symmetries when exploited for filter design. The first column indicates the filter that is obtained by applying equivariant filter design methodology with the symmetry in the second column. The third column describes the features of the state error linearization for the specific filter, whereas the rightmost column shows the linearized error dynamics. 
    For readers that do not have time to compute the linearisations themselves, detailed derivations are provided in \ifdefined\preprint appendix \ref{sec:append} \else the extended preprint~\cite[Appendix A]{Fornasier2023EquivariantSystems}. \fi }
    \begin{tblr}[
        caption = {Qualitative overview of the differences in the presented symmetries when exploited for filter design.},
    ]{
        rows = {m},
        column{1} = {2.75cm, c}, 
        column{2} = {3.35cm, c}, 
        column{3} = {4.25cm, c},
        column{4} = {5.5cm, c},
    }
        \toprule
        Filter & Symmetry group & State error linearization. 
        & ${\mathbf{A} \st \dot{\Vector{}{\varepsilon}{}} \simeq \mathbf{A}\Vector{}{\varepsilon}{}}$\\ 
        \midrule
        \ac{mekf} \cite{Lefferts1982KalmanEstimation} & {Special Orthogonal group $\grpE: \SO(3) \times \R^{12}$} & {State-dependent attitude error dynamics. State-dependent and input-dependent velocity error dynamics. Linear time-invariant position error dynamics. Linear time-invariant bias error dynamics} &
        \SetCell[]{l} {
        ${\dot{\Vector{}{\varepsilon}{R}} \simeq -\hatRot{}{}\Vector{}{\varepsilon}{b_{\omega}} + \calO\left(\Vector{}{\varepsilon}{}^2\right)}$,\\
        ${\dot{\Vector{}{\varepsilon}{v}} \simeq -\left(\hatRot{}{}\left(\Vector{}{a}{} - \hatVector{}{b}{a}\right)\right)^{\wedge}\Vector{}{\varepsilon}{R} - \hatRot{}{}\Vector{}{\varepsilon}{b_{a}} + \calO\left(\Vector{}{\varepsilon}{}^2\right)}$,\\
        ${\dot{\Vector{}{\varepsilon}{p}} \simeq \Vector{}{\varepsilon}{v}}$,\\
        ${\dot{\Vector{}{\varepsilon}{b}} = \Vector{}{0}{}}$.
        } \\
        Imperfect-\ac{iekf} \cite{7523335} & {Extended Special Euclidean group $\grpF: \SE_{2}(3) \times \R^{6}$} & {State-dependent attitude, position and velocity error dynamics. Linear time-invariant bias error dynamics} &
        \SetCell[]{l} {
        ${\dot{\Vector{}{\varepsilon}{R}} \simeq -\hatRot{}{}\Vector{}{\varepsilon}{b_{\omega}} + \calO\left(\Vector{}{\varepsilon}{}^2\right)}$,\\
        ${\dot{\Vector{}{\varepsilon}{v}} \simeq \Vector{}{g}{}^{\wedge}\Vector{}{\varepsilon}{R} - \hatVector{}{v}{}^{\wedge}\hatRot{}{}\Vector{}{\varepsilon}{b_{\omega}} - \hatRot{}{}\Vector{}{\varepsilon}{b_{a}} + \calO\left(\Vector{}{\varepsilon}{}^2\right)}$,\\
        ${\dot{\Vector{}{\varepsilon}{p}} \simeq \Vector{}{\varepsilon}{v} - \hatVector{}{p}{}^{\wedge}\hatRot{}{}\Vector{}{\varepsilon}{b_{\omega}} + \calO\left(\Vector{}{\varepsilon}{}^2\right)}$,\\
        ${\dot{\Vector{}{\varepsilon}{b}} = \Vector{}{0}{}}$.
        } \\
        \ac{tfgiekf} \cite{Barrau2022TheProblems} & Two-Frames group $\grpC:\; \SO(3) \ltimes (\R^{6} \oplus \R^{6})$ & {Linear time-invariant attitude error dynamics. State-dependent velocity and position error dynamics. State-dependent and input-dependent bias error dynamics} &
        \SetCell[]{l} {
        ${\dot{\Vector{}{\varepsilon}{R}} \simeq -\Vector{}{\varepsilon}{b_{\omega}}}$,\\
        ${\dot{\Vector{}{\varepsilon}{v}} \simeq \Vector{}{g}{}^{\wedge}\Vector{}{\varepsilon}{R} - \hat{\Vector{}{v}{}}^{\wedge}\Vector{}{\varepsilon}{b_{\omega}} - \Vector{}{\varepsilon}{b_{a}} + \calO\left(\Vector{}{\varepsilon}{}^2\right)}$,\\
        ${\dot{\Vector{}{\varepsilon}{p}} \simeq \Vector{}{\varepsilon}{v} - \hat{\Vector{}{p}{}}^{\wedge}\Vector{}{\varepsilon}{b_{\omega}} + \calO\left(\Vector{}{\varepsilon}{}^2\right)}$,\\
        ${\dot{\Vector{}{\varepsilon}{b_{\omega}}} \simeq \left(\hatRot{}{}\left(\omega - \hatVector{}{b}{\omega}\right)\right)^{\wedge}\Vector{}{\varepsilon}{b_{\omega}} + \calO\left(\Vector{}{\varepsilon}{}^2\right)}$,\\
        ${\dot{\Vector{}{\varepsilon}{b_{a}}} \simeq \left(\hatRot{}{}\left(\omega - \hatVector{}{b}{\omega}\right)\right)^{\wedge}\Vector{}{\varepsilon}{b_{a}} + \calO\left(\Vector{}{\varepsilon}{}^2\right)}$.
        } \\
        TG-\ac{eqf} \cite{Fornasier2022EquivariantBiases} & Tangent group $\grpD:\; \SE_2(3) \ltimes \gothse_2(3)$ & {Linear time-invariant attitude, velocity and position error dynamics. State-dependent and input-dependent bias error dynamics} &
        \SetCell[]{l} {
        ${\dot{\Vector{}{\varepsilon}{R}} \simeq \Vector{}{\varepsilon}{b_{\omega}}}$,\\
        ${\dot{\Vector{}{\varepsilon}{v}} \simeq \Vector{}{g}{}^{\wedge}\Vector{}{\varepsilon}{R} + \Vector{}{\varepsilon}{b_{a}}}$,\\
        ${\dot{\Vector{}{\varepsilon}{p}} \simeq \Vector{}{\varepsilon}{v} + \Vector{}{\varepsilon}{b_{\nu}}}$,\\
        ${\dot{\Vector{}{\varepsilon}{b}} \simeq \adMsym{\left(\mathring{\Vector{}{w}{}}^{\wedge} + \mathbf{G}\right)}\Vector{}{\varepsilon}{b} + \calO\left(\Vector{}{\varepsilon}{}^2\right)}$.
        } \\
        DP-\ac{eqf} & Direct Position group\footnotemark $\grpA:\; \HG(3) \ltimes \gothhg(3) \times \R^{3}$ & {Linear time-invariant attitude and velocity error dynamics. State-dependent and input-dependent position and bias error dynamics} &
        \SetCell[]{l} {
        ${\dot{\Vector{}{\varepsilon}{R}} \simeq \Vector{}{\varepsilon}{b_{\omega}}}$,\\
        ${\dot{\Vector{}{\varepsilon}{v}} \simeq \Vector{}{g}{}^{\wedge}\Vector{}{\varepsilon}{R} + \Vector{}{\varepsilon}{b_{a}}}$,\\
        ${\dot{\Vector{}{\varepsilon}{p}} \simeq \Vector{}{\varepsilon}{v} - \mathring{\Vector{}{\nu}{}}\Vector{}{\varepsilon}{R} + \calO\left(\Vector{}{\varepsilon}{}^2\right)}$,\\
        ${\dot{\Vector{}{\varepsilon}{b}} \simeq \adMsym{\left(\mathring{\Vector{}{w}{}}^{\wedge} + \mathbf{G}\right)}\Vector{}{\varepsilon}{b} + \calO\left(\Vector{}{\varepsilon}{}^2\right)}$.
        } \\ 
        SD-\ac{eqf} & Semi-Direct Bias group
        $\grpB:\; \SE_2(3) \ltimes \gothse(3)$ & {Linear time-invariant attitude and velocity error dynamics. State-dependent position error dynamics. State-dependent and input-dependent bias error dynamics} &
        \SetCell[]{l} {
        ${\dot{\Vector{}{\varepsilon}{R}} \simeq \Vector{}{\varepsilon}{b_{\omega}}}$,\\
        ${\dot{\Vector{}{\varepsilon}{v}} \simeq \Vector{}{g}{}^{\wedge}\Vector{}{\varepsilon}{R} + \Vector{}{\varepsilon}{b_{a}}}$,\\
        ${\dot{\Vector{}{\varepsilon}{p}} \simeq \Vector{}{\varepsilon}{v} + \hatVector{}{p}{}^{\wedge}\Vector{}{\varepsilon}{b_{\omega}} + \calO\left(\Vector{}{\varepsilon}{}^2\right)}$,\\
        ${\dot{\Vector{}{\varepsilon}{b}} \simeq \adMsym{\left(\mathring{\Vector{}{w}{}}^{\wedge} + \mathbf{G}\right)}\Vector{}{\varepsilon}{b} + \calO\left(\Vector{}{\varepsilon}{}^2\right)}$.
        } \\
        \bottomrule
    \end{tblr}\label{tab:symmetries_overview}
\end{table*}

\section{INS Symmetries}\label{sec:sym}
Starting with \tabref{symmetries_overview}, we show the relation between \ac{ins} filters and symmetry group, as well as the differences in the state error linearization of filters built upon those symmetries. 
In Sec.~\ref{sec:SO3_symmetry}, \ref{sec:SE23_symmetry} and \ref{sec:tf_symmetry}, we discuss the symmetry groups that lead to the design of \aclp{eqf} equivalent to the widely-known \ac{mekf}, \ac{iekf}, and the recently published TFG-\ac{iekf}. 
In \secref{SE23_se23_symmetry} we briefly recall the tangent group recently introduced and exploited for \ac{ins} in our prior work~\cite{Fornasier2022EquivariantBiases, Fornasier2022OvercomingCalibration}. 
In Sec.~\ref{sec:se3_R3_symmetry} and \ref{sec:se23_symmetry}, we introduce two new symmetry groups for biased inertial navigation Systems. 
These groups are based on the semi-direct product and aim to address the over-parametrization of bias states introduced in our prior work~\cite{Fornasier2022EquivariantBiases}.

%

\subsection{Special Orthogonal group ${\grpE: \SO(3) \times \R^{12}}$}\label{sec:SO3_symmetry}
Lie group theory was first applied to navigation systems to overcome the limitation and the singularities of using Euler angles as the parameterization of the attitude of a rigid body. 
Originally formulated on the quaternion group, the modern approach directly models attitude on the Special Orthogonal group $\SO(3)$.

\sloppy Define the symmetry group ${\grpE \coloneqq \SO(3) \times \R^{12}}$, and let ${X = \left(A, a, b, \alpha, \beta\right) \in \grpE}$, where ${A \in \SO(3)}$, ${a, b, \alpha, \beta \in \R^3}$.
Let ${X = \left(A_X, a_X, b_X, \alpha_X, \beta_X\right)}$, ${Y = \left(A_Y, a_Y, b_Y, \alpha_Y, \beta_Y\right)}$ be two elements of the symmetry group, then the group product is written  ${XY = \left(A_XA_Y, a_X + a_Y, b_X + b_Y, \alpha_X + \alpha_Y, \beta_X + \beta_Y\right)}$.
The inverse of an element $X$ is given by ${X^{-1} = \left(A^\top, -a, -b, -\alpha, -\beta\right)}$.

\begin{lem}
Define ${\phi \AtoB{\grpE \times \calM}{\calM}}$ as
\begin{equation}
    \phi\left(X, \xi\right) \coloneqq \left(\Rot{}{}A, \Vector{}{v}{} + a, \Vector{}{p}{} + b, \Vector{}{b}{\bm{\omega}} + \alpha, \Vector{}{b}{a} + \beta\right) \in \calM .
\end{equation}
Then, $\phi$ is a transitive right group action of $\grpE$ on $\calM$.
\end{lem}

The existence of a transitive group action of the symmetry group $\grpE$ on the state space $\calM$ guarantees the existence of a lift~\cite{Mahony2020EquivariantDesign}.

\begin{thm}
Define the map ${\Lambda \AtoB{\calM \times \vecL}{\gothGrpE}}$ by
\begin{equation*}
    \Lambda\left(\xi, u\right) \coloneqq \left(\Lambda_1\left(\xi, u\right), \cdots, \Lambda_5\left(\xi, u\right)\right).
\end{equation*}
where ${\Lambda_1 \AtoB{\calM \times \vecL}{\gothso(3)}}$, and ${\Lambda_2, \cdots, \Lambda_5 \AtoB{\calM \times \vecL}{\R^{3}}}$ are given by
\begin{align}
    &\Lambda_1\left(\xi, u\right) \coloneqq \left(\Vector{I}{\bm{\omega}}{} - \Vector{I}{b}{\bm{\omega}}\right)^{\wedge} ,\\
    &\Lambda_2\left(\xi, u\right) \coloneqq \Rot{G}{I}\left(\Vector{I}{a}{} - \Vector{I}{b}{a}\right) + \Vector{G}{g}{} ,\\
    &\Lambda_3\left(\xi, u\right) \coloneqq \Vector{G}{v}{I} ,\\
    &\Lambda_4\left(\xi, u\right) \coloneqq \Vector{I}{\tau}{\bm{\omega}} ,\\
    &\Lambda_4\left(\xi, u\right) \coloneqq \Vector{I}{\tau}{a} .
\end{align}
Then, the ${\Lambda}$ is a lift for the system in \equref{bins} with respect to the symmetry group ${\grpE \coloneqq \SO(3) \times \R^{12}}$.
\end{thm}

In the 
\ifdefined\preprint
appendix \ref{sec:append},
\else
extended preprint~\cite[Appendix A]{Fornasier2023EquivariantSystems},
\fi 
it is shown that an \ac{eqf} designed using this symmetry results in the well-known \ac{mekf}~\cite{Lefferts1982KalmanEstimation}.

\subsection{Extended Special Euclidean group ${\grpF: \SE_2(3) \times \R^{6}}$}\label{sec:SE23_symmetry}
Using the extended pose $\SE_2(3)$ group to model the navigation states of the INS problem is one of the major developments in INS filtering in the last 10 years. 

Define ${\xi = \left(\Pose{}{}, \Vector{}{b}{}\right) \in \calM \coloneqq \torSE_2(3) \times \R^{6}}$ to be the state space of the system. ${\Pose{}{} = \left(\Rot{}{}, \Vector{}{v}{}, \Vector{}{p}{}\right) \in \torSE_2(3)}$ is the extended pose~\cite{Brossard2021AssociatingEarth}, which includes the orientation the, velocity and the position of the rigid body, whereas ${\Vector{}{b}{} = \left(\Vector{}{b}{\bm{\omega}}, \Vector{}{b}{\bm{a}}\right) \in \R^{6}}$ denotes the \ac{imu} biases.
Let ${u = \left(\Vector{}{w}{}, \bm{\tau}\right) \in \vecL \subseteq \R^{12}}$ denote the system input, where ${\Vector{}{w}{} = \left(\Vector{}{\bm{\omega}}{}, \Vector{}{a}{}\right) \in \R^{6}}$ denotes the input given by the \ac{imu} readings, and ${\bm{\tau} = \left(\bm{\tau_{\omega}}, \bm{\tau_{a}}\right) \in \R^{6}}$ denotes the input for the \ac{imu} biases.
\footnotetext{The homogeneous Galilean group $\HG(3)$ is isomorphic to $\SE(3)$ but acts on attitude and velocity rather than attitude and position.}
Define the matrices
\begin{equation*}
    \begin{tblr}{ll}
    \mathbf{G} = (\ubarVector{}{\lbarVector{}{g}{}}{})^{\wedge} \in \gothse_2(3), &
    \SetCell[r=3]{m}{
    \mathbf{N} = \begin{bmatrix}
    \mathbf{0}_{3\times 3} & \mathbf{0}_{3\times 1} & \mathbf{0}_{3\times 1}\\
    \mathbf{0}_{1\times 3} & 0 & 1\\
    \mathbf{0}_{1\times 3} & 0 & 0\\
    \end{bmatrix} \in \R^{5 \times 5}.
    } \\
    \mathbf{B} = (\lbarVector{}{b}{})^{\wedge} \in \gothse_2(3), &\\
    \mathbf{W} = (\lbarVector{}{w}{})^{\wedge} \in \gothse_2(3), &\\
    \end{tblr}
\end{equation*}
Then, the system in \equref{bins} may then be written as
\begin{subequations}\label{eq:bins_se23}
    \begin{align}
        &\dot{\Pose{}{}} = \Pose{}{}\left(\mathbf{W} - \mathbf{B} + \mathbf{N}\right) + \left(\mathbf{G} - \mathbf{N}\right)\Pose{}{} ,\\
        &\dotVector{}{b}{} =  \bm{\tau} .
    \end{align}
\end{subequations}

Define the symmetry group ${\grpF \coloneqq \SE_2(3) \times \R^{6}}$, and let ${X = \left(C, \gamma\right) \in \grpF}$, where ${C = \left(A, a, b\right)\in \SE_2(3)}$, $A \in \SO(3)$, $a, b \in \R^3$, ${\gamma \in \R^6}$.
Let ${X = \left(C_X, \gamma_X\right)}$, ${Y = \left(C_Y, \gamma_Y\right)}$ be two elements of the symmetry group, then the group product is written  ${XY = \left(C_XC_Y, \gamma_X + \gamma_Y\right)}$.
The inverse of an element $X$ is given by ${X^{-1} = \left(C^{-1}, -\gamma\right)}$.

\begin{lem}
Define ${\phi \AtoB{\grpF \times \calM}{\calM}}$ as
\begin{equation}
    \phi\left(X, \xi\right) \coloneqq \left(\Pose{}{}C, \Vector{}{b}{} + \gamma\right) \in \calM .
\end{equation}
Then, $\phi$ is a transitive right group action of $\grpF$ on $\calM$.
\end{lem}

\begin{thm}
Define the map ${\Lambda \AtoB{\calM \times \vecL}{\gothGrpF}}$ by
\begin{equation*}
    \Lambda\left(\xi, u\right) \coloneqq \left(\Lambda_1\left(\xi, u\right), \Lambda_2\left(\xi, u\right)\right),
\end{equation*}
where ${\Lambda_1 \AtoB{\calM \times \vecL}{\gothse_2(3)}}$, and ${\Lambda_2 \AtoB{\calM \times \vecL}{\R^{6}}}$ are given by
\begin{align}
    &\Lambda_1\left(\xi, u\right) \coloneqq \left(\mathbf{W} - \mathbf{B} + \mathbf{N}\right) + \Pose{}{}^{-1}\left(\mathbf{G} - \mathbf{N}\right)\Pose{}{} ,\\
    &\Lambda_2\left(\xi, u\right) \coloneqq \bm{\tau} .
\end{align}
Then, ${\Lambda}$ is a lift for the system in \equref{bins_se23} with respect to the symmetry group ${\grpF \coloneqq \SE_2(3) \times \R^{6}}$.
\end{thm}

Applying the EqF filter design methodology to this symmetry leads to the Imperfect-\ac{iekf}~\cite{7523335, barrau:tel-01247723}. 
Note that ignoring the bias and considering only the navigation states is the original IEKF filter~\cite{7523335}.
The imperfect term comes from breaking the group-affine symmetry of the navigation states by adding the direct product terms for the bias. 

\subsection{Two-Frames group: ${\grpC:\; \SO(3) \ltimes (\R^{6} \oplus \R^{6})}$}\label{sec:tf_symmetry}
The recently published \acl{tfgiekf}~\cite{Barrau2022TheProblems} is one approach to address the theoretical issue in the imperfect IEKF for INS where the bias terms are not part of the symmetry structure. 

Consider the system in \equref{bins_se23}. Define the symmetry group ${\grpC \coloneqq \SO(3) \ltimes (\R^{6} \oplus \R^{6})}$, where ${\SO(3)}$ acts on two vector spaces of 6 dimensions each defined with respect to two different frames of references. Let ${X = \left(C,\gamma\right) \in \grpC}$, with ${C = \left(A, \left(a, b\right)\right) \in \SE_2(3) \coloneqq \SO(3) \ltimes \R^{6}}$ such that ${A \in \SO(3),\; \left(a,b\right) \in \R^{6}}$. Let, $* \AtoB{\SO(3) \times \R^{3N}}{\R^{3N}}$ be the rotation term introduced in~\cite{Barrau2022TheProblems}, such that ${\forall\, A \in \SO(3)}$ and ${x = \left(x_1,\cdots,x_N\right) \in \R^{3N}}$, ${A*x = \left(Ax_1,\cdots,Ax_N\right)}$.
Define the group product  ${XY = \left(C_XC_Y, \gamma_X + A_X * \gamma_Y\right)}$
The inverse element of the symmetry group writes ${X^{-1} = \left(C^{-1},-A^{T}*\gamma\right) \in \grpC}$.

\begin{lem}
Define ${\phi \AtoB{\grpC \times \calM}{\calM}}$ as
\begin{equation}\label{eq:phi_tf}
    \phi\left(X, \xi\right) \coloneqq \left(\Pose{}{}C, A^{T}*\left(\Vector{}{b}{} - \gamma\right)\right) \in \calM .
\end{equation}
Then, $\phi$ is a transitive right group action of $\grpC$ on $\calM$.
\end{lem}

\begin{thm}
Define ${\Lambda_1 \AtoB{\calM \times \vecL} \se_2(3)}$ as
\begin{equation}\label{eq:lift1_tf}
    \Lambda_1\left(\xi, u\right) \coloneqq \left(\mathbf{W} - \mathbf{B} + \mathbf{N}\right) + \Pose{}{}^{-1}\left(\mathbf{G} - \mathbf{N}\right)\Pose{}{} ,
\end{equation}
define ${\Lambda_2 \AtoB{\calM \times \vecL} \R^{6}}$ as
\begin{equation}\label{eq:lift2_tf}
    \Lambda_2\left(\xi, u\right) \coloneqq \left(\Vector{}{b}{\bm{\omega}}^{\wedge}\left(\bm{\omega} - \Vector{}{b}{\bm{\omega}}\right) - \bm{\tau_{\omega}} ,\; \Vector{}{b}{\bm{a}}^{\wedge}\left(\bm{\omega} - \Vector{}{b}{\bm{\omega}}\right) - \bm{\tau_a}\right) .
\end{equation}
Then, the map ${\Lambda\left(\xi, u\right) = \left(\Lambda_1\left(\xi, u\right), \Lambda_2\left(\xi, u\right)\right)}$ is a lift for the system in \equref{bins_se23} with respect to the symmetry group ${\grpC \coloneqq \SO(3) \ltimes (\R^{6} \oplus \R^{6})}$.
\end{thm}

In the 
\ifdefined\preprint
appendix \ref{sec:append},
\else
extended preprint~\cite[Appendix A]{Fornasier2023EquivariantSystems},
\fi 
it is shown that designing and EqF based on this symmetry leads to the recently published \ac{tfgiekf}~\cite{Barrau2022TheProblems}.

\subsection{Tangent group ${\grpD:\; \SE_2(3) \ltimes \gothse_2(3)}$}\label{sec:SE23_se23_symmetry}
Recent work \cite{ng2019attitude, Ng2020PoseKinematics} considered symmetries and EqF filter design on the tangent group $\tT \grpG$ of a general Lie-group. 
Since bias states are closely related to velocities, these ideas can easily be extended symmetries for bias states~\cite{Fornasier2022EquivariantBiases}. 

Define ${\xi = \left(\Pose{}{}, \Vector{}{b}{}\right) \in \calM \coloneqq \torSE_2(3) \times \R^{9}}$ to be the state space of the system. ${\Pose{}{} \in \torSE_2(3)}$ represents the extended pose, whereas ${\Vector{}{b}{} = \left(\Vector{}{b}{\bm{\omega}}, \Vector{}{b}{\bm{a}}, \Vector{}{b}{\bm{\nu}}\right) \in \R^{9}}$ represents the \ac{imu} biases, and an additional virtual bias $\Vector{}{b}{\bm{\nu}}$.
Let ${u = \left(\Vector{}{w}{}, \bm{\tau}\right) \in \vecL \subseteq \R^{18}}$ denote the system input, where ${\Vector{}{w}{} = \left(\Vector{}{\bm{\omega}}{}, \Vector{}{a}{}, \Vector{}{\nu}{}\right) \in \R^{9}}$ denotes the input given by the \ac{imu} readings, and an additional virtual input $\Vector{}{\nu}{}$. Note that we can set $\Vector{}{\nu}{} = \Vector{}{b}{\bm{\nu}}$ such that the original dynamics in \equref{bins} are recovered. ${\bm{\tau} = \left(\bm{\tau_{\omega}}, \bm{\tau_{a}}, \bm{\tau_{\nu}}\right) \in \R^{9}}$ denotes the input for the \ac{imu} biases.
Define the matrices
\begin{equation*}
    \begin{tblr}{ll}
    \mathbf{G} = (\ubarVector{}{\lbarVector{}{g}{}}{})^{\wedge} \in \gothse_2(3), &
    \SetCell[r=3]{m}{
    \mathbf{N} = \begin{bmatrix}
    \mathbf{0}_{3\times 3} & \mathbf{0}_{3\times 1} & \mathbf{0}_{3\times 1}\\
    \mathbf{0}_{1\times 3} & 0 & 1\\
    \mathbf{0}_{1\times 3} & 0 & 0\\
    \end{bmatrix} \in \R^{5 \times 5}.
    } \\
    \mathbf{B} = \Vector{}{b}{}^{\wedge} \in \gothse_2(3), &\\
    \mathbf{W} = \Vector{}{w}{}^{\wedge} \in \gothse_2(3), &\\
    \end{tblr}
\end{equation*}

With these newly defined matrices, the system in \equref{bins} may then be written in compact form as in \equref{bins_se23}. Note, however, that the matrices $\mathbf{B}$ and $\mathbf{W}$ are different than those in \equref{bins_se23}.

Define the symmetry group ${\grpD \coloneqq \SE_2(3) \ltimes \gothse_2(3)}$, and let ${X = \left(C, \gamma\right) \in \grpD}$, where ${C \in \SE_2(3)}$, ${\gamma \in \gothse_2(3)}$.
Let ${X = \left(C_X, \gamma_X\right)}$, ${Y = \left(C_Y, \gamma_Y\right)}$ be two elements of the symmetry group, then the group product is written  ${XY = \left(C_XC_Y, \gamma_X + \Adsym{C_X}{\gamma_Y}\right)}$.
The inverse of an element $X$ is given by ${X^{-1} = \left(C^{-1}, -\Adsym{C^{-1}}{\gamma}\right)}$.

\begin{lem}
Define ${\phi \AtoB{\grpD \times \calM}{\calM}}$ as
\begin{equation}\label{eq:phi_se23_tg}
    \phi\left(X, \xi\right) \coloneqq \left(\Pose{}{}C, \AdMsym{C^{-1}}\left(\Vector{}{b}{} - \gamma^{\vee}\right)\right) \in \calM .
\end{equation}
Then, $\phi$ is a transitive right group action of $\grpD$ on $\calM$.
\end{lem}
From here, we derive a compatible action of the symmetry group ${\grpD}$ on the input space ${\vecL}$ and derive the lift ${\Lambda}$ via constructive design as described in~\cite{Mahony2020EquivariantDesign, VanGoor2020EquivariantSpaces}.
\begin{lem}
Define ${\psi \AtoB{\grpD \times \vecL}{\vecL}}$ as
\begin{equation}\label{eq:psi_se23}
    \psi\left(X, u\right) \coloneqq \left(\AdMsym{C^{-1}}\left(\Vector{}{w}{} - \gamma^{\vee}\right) + \Omega^{\vee}\left(C^{-1}\right), \AdMsym{C^{-1}}\Vector{}{\tau}{}\right) \in \vecL .
\end{equation}
Then, $\psi$ is a right group action of $\grpD$ on $\vecL$.
\end{lem}
The system in \equref{bins_se23} is equivariant under the actions $\phi$ in \equref{phi_se23_tg} and $\psi$ in \equref{psi_se23}. The existence of a transitive group action of the symmetry group $\grpD$ on the state space $\calM$ and the equivariance of the system guarantees the existence of an equivariant lift~\cite{Mahony2020EquivariantDesign}.
\begin{thm}
Define the map ${\Lambda \AtoB{\calM \times \vecL}{\gothGrpD}}$ by
\begin{equation*}
    \Lambda\left(\xi, u\right) \coloneqq \left(\Lambda_1\left(\xi, u\right), \Lambda_2\left(\xi, u\right)\right),
\end{equation*}
where ${\Lambda_1 \AtoB{\calM \times \vecL}{\gothse_2(3)}}$, and ${\Lambda_2 \AtoB{\calM \times \vecL}{\gothse_2(3)}}$ are given by
\begin{align}
    &\Lambda_1\left(\xi, u\right) \coloneqq \left(\mathbf{W} - \mathbf{B} + \mathbf{N}\right) + \Pose{}{}^{-1}\left(\mathbf{G} - \mathbf{N}\right)\Pose{}{} ,\\
    &\Lambda_2\left(\xi, u\right) \coloneqq \adsym{\Vector{}{b}{}}{\Lambda_1\left(\xi, u\right)} - \bm{\tau}^{\wedge}.
\end{align}
Then, ${\Lambda}$ is an equivaraint lift for the system in \equref{bins_se23} with respect to the symmetry group ${\grpD \coloneqq \SE_2(3) \times \gothse_2(3)}$.
\end{thm}

This approach requires the introduction of a new state $\Vector{}{b}{\bm{\nu}}$ in order to apply the full $\se_2(3)$ semi-direct symmetry on the bias states. 
This new state is entirely virtual, it does not exist in the real system. 
Since introducing an entirely virtual state just for the sake of the symmetry appears questionable, it is of interest to consider alternative symmetries that try to preserve the semi-direct group structure that models bias interaction, but doesn't require the additional bias filter state.

\subsection{Direct Position group ${\grpA:\; \HG(3) \ltimes \gothhg(3) \times \R^{3}}$}\label{sec:se3_R3_symmetry}
In this subsection, we investigate a symmetry that does not require the over-parametrization of the state introduced in~\cite{Fornasier2022EquivariantBiases} given by the addition of a velocity bias state. Specifically, we achieve this by considering a semi-direct product symmetry only on the homogeneous Galilean structure of the state space and an Euclidean symmetry for the position state.

We introduce the term $\HG(3)$ for the \emph{homogeneous Galilean} group.  
This corresponds to extended pose transformations $\SE_2(3)$ where the spatial translation is zero. 
That is the symmetry acts on rotation and velocity only with the semi-direct product induced by the $\SE_2(3)$ geometry. 
The homogeneous Galilean group is isomorphic to $\SE(3)$ in structure, however, since $\SE(3)$ is synonymous with pose transformation we use the $\HG(3)$ notation to avoid confusion. 

The first step towards these goals is to introduce a virtual input ${\bm{\nu}}$ and rewrite \equref{bins_p} as ${\dotVector{}{p}{} = \Rot{}{}\bm{\nu} + \Vector{}{v}{}}$. Note that the input $\bm{\nu}$ can be set to zero to recover the original dynamics in~\equref{bins}.

Define ${\xi = \left(\Pose{}{}, \Vector{}{p}{}, \Vector{}{b}{}\right) \in \calM \coloneqq \calHG(3) \times \R^{3} \times \R^{6}}$ to be the state space of the system, where ${\Pose{}{} = \left(\Rot{}{}, \Vector{}{v}{}\right) \in \calHG(3)}$ includes the orientation and the velocity of the rigid body.
Let ${u = \left(\Vector{}{w}{}, \bm{\nu}, \bm{\tau}\right) \in \vecL \subseteq \R^{15}}$ denote the system input.
Define the matrices
\begin{equation*}
    \mathbf{G} = (\ubarVector{}{g}{})^{\wedge} \in \gothse(3), \quad
    \mathbf{B} = \Vector{}{b}{}^{\wedge} \in \gothse(3), \quad
    \mathbf{W} = \Vector{}{w}{}^{\wedge} \in \gothse(3).
\end{equation*}
Then, the system in \equref{bins} may then be written as
\begin{subequations}\label{eq:bins_se3_R3}
    \begin{align}
        &\dot{\Pose{}{}} = \Pose{}{}\left(\mathbf{W} - \mathbf{B}\right) + \mathbf{G}\Pose{}{} ,\\
        &\dotVector{}{p}{} = \Rot{}{}\bm{\nu} + \Vector{}{v}{} ,\label{eq:bins_se3_R3_p}\\
        &\dotVector{}{b}{} =  \bm{\tau} .
    \end{align}
\end{subequations}

Define the symmetry group ${\grpA \coloneqq \HG(3) \ltimes \gothse(3) \times \R^{3}}$, and let ${X = \left(B,\beta,c\right) \in \grpA}$ with ${B = \left(A, a\right) \in \HG(3)}$ such that ${A \in \SO(3),\; a \in \R^{3}}$.
Let $X = \left(B_X,\beta_X,c_X\right), Y = \left(B_Y,\beta_Y,c_Y\right) \in \grpA$, the group product is written  ${XY = \left(B_XB_Y, \beta_X + \Adsym{B_X}{\beta_Y}, c_X + c_Y\right)}$.
The inverse of an element $X \in \grpA$ is given by ${X^{-1} = \left(B^{-1},-\Adsym{B^{-1}}{\beta},-c\right) \in \grpA}$. 

\begin{lem}
Define ${\phi \AtoB{\grpA \times \calM}{\calM}}$ as
\begin{equation}\label{eq:phi_se3_R3}
    \phi\left(X, \xi\right) \coloneqq \left(\Pose{}{}B, \AdMsym{B^{-1}}\left(\Vector{}{b}{} - \beta^{\vee}\right), \Vector{}{p}{} + c\right) \in \calM .
\end{equation}
Then, $\phi$ is a transitive right group action of $\grpA$ on $\calM$.
\end{lem}
We derive a compatible action of the symmetry group ${\grpA}$ on the input space ${\vecL}$.
\begin{lem}
Define ${\psi \AtoB{\grpA \times \vecL}{\vecL}}$ as
\begin{equation}\label{eq:psi_se3_R3}
    \psi\left(X, u\right) \coloneqq \left(\AdMsym{B^{-1}}\left(\Vector{}{w}{} - \beta^{\vee}\right), A^T\left(\bm{\nu} - a\right), \AdMsym{B^{-1}}\Vector{}{\tau}{}\right) \in \vecL .
\end{equation}
Then, $\psi$ is a right group action of $\grpA$ on $\vecL$.
\end{lem}
The system in \equref{bins_se3_R3} is equivariant under the actions $\phi$ in \equref{phi_se3_R3} and $\psi$ in \equref{psi_se3_R3}. Therefore, the existence  of an equivariant lift is guaranteed.
\begin{thm}
Define the map ${\Lambda \AtoB{\calM \times \vecL}{\gothGrpA}}$ by
\begin{align*}
    \Lambda\left(\xi, u\right) &\coloneqq \left(\Lambda_1\left(\xi, u\right), \Lambda_2\left(\xi, u\right), \Lambda_3\left(\xi, u\right)\right),
\end{align*}
where ${\Lambda_1 \AtoB{\calM \times \vecL} \gothhg(3)}$, ${\Lambda_2 \AtoB{\calM \times \vecL} \se(3)}$, and ${\Lambda_3 \AtoB{\calM \times \vecL} \R^{3}}$ are given by
\begin{align}
    \label{eq:lift1_se3_R3_1}
    \Lambda_1\left(\xi, u\right) &\coloneqq \left(\mathbf{W} - \mathbf{B}\right) + \Pose{}{}^{-1}\mathbf{G}\Pose{}{} , \\
    \label{eq:lift2_se3_R3_2}
    \Lambda_2\left(\xi, u\right) &\coloneqq \adsym{\Vector{}{b}{}^{\wedge}}{\Lambda_1\left(\xi, u\right)} - \bm{\tau}^{\wedge} , \\
    \label{eq:lift3_se3_R3_3}
    \Lambda_3\left(\xi, u\right) &\coloneqq \Rot{}{}\bm{\nu} + v .
\end{align}
Then, the ${\Lambda}$ is an equivariant lift for the system in \equref{bins_se3_R3} with respect to the symmetry group ${\grpA \coloneqq \HG(3) \ltimes \gothhg(3) \times \R^{3}}$.
\end{thm}

The symmetry proposed in this subsection allows for a minimal state parametrization (i.e. absence of over-parametrization of the state with additional state variables ).
However, the construction comes at the cost of separating the position state from the geometric $\SE_2(3)$ structure and modeling it as a direct product linear space.

\subsection{Semi-Direct Bias group: ${\grpB:\; \SE_2(3) \ltimes \gothse(3)}$}\label{sec:se23_symmetry}
In this subsection, we investigate a symmetry that maintains a minimal state representation (not requiring the introduction of an additional velocity bias state) while keeping the position state within the geometric structure given by ${\SE_2(3)}$.

Consider the system in \equref{bins_se23}. Define the symmetry group ${\grpB \coloneqq \SE_2(3) \ltimes \gothse(3)}$ with group product ${XY = \left(C_XC_Y, \gamma_X + \Adsym{C_X}{\gamma_Y}\right)}$ 
for $X = \left(C_X,\gamma_X\right), Y = \left(C_Y,\gamma_Y\right) \in \grpB$. 
Here, for ${X = \left(C,\gamma\right) \in \grpB}$ one has ${C = \left(A, a, b\right) = \left(B, b\right)\in \SE_2(3)}$ such that ${A \in \SO(3),\; a,b \in \R^{3}}$, and ${B = \left(A, a\right) \in \HG(3)}$. 
The element $C \in \SE_2(3)$ in its matrix representation is written
\begin{equation*}
    C = \begin{bNiceArray}{w{c}{0.75cm}w{c}{0.45cm}:w{c}{0.45cm}}[margin]
        A & a & b\\
        \mathbf{0}_{1\times 3} & 1 & 0\Bstrut\\
        \hdottedline
        \mathbf{0}_{1\times 3} & 0 & 1\Tstrut
    \end{bNiceArray} =
    \begin{bNiceArray}{w{c}{0.75cm}w{c}{0.45cm}:w{c}{0.45cm}}[margin]
        \Block{2-2}{B} & & b\\
        & & 0\Bstrut\\
        \hdottedline
        \mathbf{0}_{1\times 3} & 0 & 1\Tstrut
    \end{bNiceArray} \in \SE_2(3) .
\end{equation*}
The inverse element is written 
\[
{X^{-1} = \left(C^{-1},-\Adsym{B^{-1}}{\gamma}\right) \in \grpB}.
\]

\begin{lem}
Define ${\phi \AtoB{\grpB \times \calM}{\calM}}$ as
\begin{equation}\label{eq:phi_se23_sdb}
    \phi\left(X, \xi\right) \coloneqq \left(\Pose{}{}C, \AdMsym{B^{-1}}\left(\Vector{}{b}{} - \gamma^{\vee}\right)\right) \in \calM .
\end{equation}
Then, $\phi$ is a transitive right group action of $\grpB$ on $\calM$.
\end{lem}

\begin{thm}
Define ${\Lambda_1 \AtoB{\calM \times \vecL} \se_2(3)}$ as
\begin{equation}\label{eq:lift1_se23}
    \Lambda_1\left(\xi, u\right) \coloneqq \left(\mathbf{W} - \mathbf{B} + \mathbf{N}\right) + \Pose{}{}^{-1}\left(\mathbf{G} - \mathbf{N}\right)\Pose{}{} ,
\end{equation}
define ${\Lambda_2 \AtoB{\calM \times \vecL} \se(3)}$ as
\begin{equation}\label{eq:lift2_se23}
    \Lambda_2\left(\xi, u\right) \coloneqq \adsym{\Vector{}{b}{}^{\wedge}}{\Pi\left(\Lambda_1\left(\xi, u\right)\right)} - \bm{\tau}^{\wedge} ,
\end{equation}
Then, the map ${\Lambda\left(\xi, u\right) = \left(\Lambda_1\left(\xi, u\right), \Lambda_2\left(\xi, u\right)\right)}$ is a lift for the system in \equref{bins_se23} with respect to the symmetry group ${\grpB \coloneqq \SE_2(3) \ltimes \gothse(3)}$.
\end{thm}

The symmetry proposed in this subsection is a variation of the symmetry defined in our previous work~\cite{Fornasier2022EquivariantBiases} that does not require over-parametrization of the state and additional state variables.

%% file: sections/linearisation.tex
\section{Linearization Error Analysis}\label{sec:le_analysis}

In \secref{sym}, we present different symmetry groups for the inertial navigation problem.
An indicator of the performance of an \ac{eqf} with a particular symmetry is the order of approximation error in the associated linearization of error dynamics.

For all symmetries, the origin $\mathring{\xi}\in\calM$ is chosen to be $\mathring{\xi} := \left(\mathring{\Rot{}{}}, \mathring{\Vector{}{v}{}}, \mathring{\Vector{}{p}{}}, \mathring{\Vector{}{b}{\bm{\omega}}}, \mathring{\Vector{}{b}{a}}\right) = \left(\eye_3, \Vector{}{0}{3\times1},\Vector{}{0}{3\times1},\Vector{}{0}{3\times1},\Vector{}{0}{3\times1}\right)$.
Define the local coordinate chart $\vartheta:\mathcal{U}_{\mathring{\xi}}\to\R^n$, to be 
\begin{align}\label{eq:local_chart}
    \vartheta \coloneqq (\phi_{\mathring{\xi}}\cdot\exp_\grpG)^{-1}= \log_\grpG\cdot\phi_{\mathring{\xi}}^{-1}, 
\end{align}
on a neighborhood of $\mathring{\xi}\in\calM$ such that $\log_\grpG\cdot\phi_{\mathring{\xi}}^{-1}$ is bijective.
The chart $\vartheta$ is always well-defined locally since all group actions considered are free.
The local error coordinates are defined by $\varepsilon \coloneqq \vartheta(e)$, so long as $e\coloneqq\phi(\hat{X}^{-1},\xi)$ remains in the domain of definition of $\vartheta$.

In \equref{local_chart}, $\log_\grpG$ denotes the log coordinates on the symmetry group considered. 
This map is different for each symmetry group. 
For a product Lie group $\grpG\coloneqq\grpG_1\times\grpG_2$, the logarithm is given by $\log_\grpG((A,B)) = (\log_{\grpG_1}(A), \log_{\grpG_2}(B))$.
When the product groups are Lie groups with well-known exponential maps, then the standard expressions are used \cite{chirikjian2011stochastic}. 
For the semi-direct product groups $\grpG\ltimes\gothg$ where $\gothg$ is the Lie algebra of $\grpG$, we will use a matrix realization to compute the exponential and the logarithm algebraically.

In the rightmost column of \tabref{symmetries_overview}, we present the linearization of the state error dynamics associated with each of the symmetries considered.
The linearization is expressed in terms of elements $\varepsilon = \log(E) \in \gothg$ where the element $E \in \grpG$ corresponds bijectively to the error $e \in \calM$ through the free group action. 
That is, we solve 
\[
\tD \vartheta^{-1}(\varepsilon)[\dot{\varepsilon}] \approx \ddt e = f(\vartheta^{-1}(\varepsilon),u) 
\]
for $\dot{\varepsilon}$ to first order in $\varepsilon$. 
Here $\dot{e} = f(e,u)$ is the full error dynamics expressed as a function of $e$ and the input $u$ \cite{Mahony2022ObserverEquivariance,vanGoor2022EquivariantEqF}. Finally, the filter design follows the procedure outlined in \secref{prelim} and in the authors' earlier works~\cite{vanGoor2022EquivariantEqF, Mahony2022ObserverEquivariance, Fornasier2022EquivariantBiases, Fornasier2022OvercomingCalibration}. The detailed derivation of the error linearization for each symmetry, as well as the related \aclp{eqf}, are provided in the 
\ifdefined\preprint
appendix \ref{sec:append}.
\else
extended preprint~\cite[Appendix A]{Fornasier2023EquivariantSystems}.
\fi 

Barrau and Bonnabel~\cite{7523335} developed the \ac{iekf} for the bias free \ac{ins} problem and showed that the linearization of the navigation states was exact. This was a significant improvement on the \ac{mekf} geometry, where the linearization of the navigation states is not exact, independently of the bias. 
However, this exact linearization property is lost when bias is added to the INS problem, the system is no longer group affine~\cite{barrau:tel-01247723}. 
Using a direct product geometric structure to add bias leads to the Imperfect-\ac{iekf}~\cite{barrau:tel-01247723} and introduces linearization error in the navigation state equations (cf.~\tabref{symmetries_overview}). 
The remaining filters all model coupling between bias and navigation states using semi-direct geometry of some form or other. 
The TG-\ac{eqf} is the only filter for which the linearization of the navigation state is exact.  
In this case, the linearization error is only present in the bias states. 
The DP-\ac{eqf}, SD-\ac{eqf} and \ac{tfgiekf} all have semi-direct geometric coupling between part of their navigation states and the bias states leading to exact or improved linearization where the coupling acts compatibly with the $\tT\grpG$ structure. 

%% file: sections/experiments.tex
\section{Application: position-based localization}\label{sec:appl}
In the present section, we discuss a practical application of the symmetries presented in \secref{sym}, that is \ac{uav} position-based localization.

Consider the system in \equref{bins}, and consider the output for global position measurements:
\begin{equation}\label{eq:confout}
     h\left(\xi\right) = \Vector{}{p}{} \in \R^{3} .
\end{equation}
It is straightforward to verify the ${\grpF, \grpD, \grpB, \grpC}$ symmetries do not possess output equivariance~\cite{VanGoor2020EquivariantSpaces, vanGoor2022EquivariantEqF} for global position measurements directly.

\subsection{Reformulation of position measurements as equivariant}\label{sec:reformulation}
Here, we show how position measurements can be reformulated as residual body-frame measurements imposing a nonlinear constraint~\cite{Julier2007OnConstraints}. 
The modified measurement is output equivariant with respect to a suitable group action, and the linearization methodology proposed in \cite{vanGoor2022EquivariantEqF} can be applied to generate cubic linearization error in the output. 

\begin{lem}
Let ${\Vector{}{\pi}{}}$ be a measurement of global position. Define a new measurement model ${h\left(\xi\right) \in \R^{3}}$, describing the body-referenced difference between the measured global position and the position state as follows
\begin{equation}\label{eq:confout_vec}
     h\left(\xi\right) = \Rot{}{}^T\left(\Vector{}{\pi}{} - \Vector{}{p}{}\right) \in \R^{3} .
\end{equation}
Let ${y = h\left(\xi\right) \in \calN}$ be a measurement defined according to the above model in \equref{confout_vec}, define ${\rho \AtoB{\grpG \times \R^{3}}{\R^{3}}}$ such that
\begin{equation}\label{eq:rho_posi}
    \rho_X\left(y\right) \coloneqq A^T\left(y - b\right).
\end{equation}
Then, the output defined in \equref{confout_vec} is equivariant.
\end{lem}

The noise-free value for $y$ is zero and the output innovation ${\delta\left(\rho_{\hat{X}^{-1}}\left({\Vector{}{0}{}}\right)\right) = \rho_{\hat{X}^{-1}}\left({\Vector{}{0}{}}\right) - \Vector{}{\pi}{}}$ measures the mismatch of the observer state in reconstructing the true state up to noise in the raw measurement ${\Vector{}{\pi}{}}$. 



\subsection{Experimental evaluation}
We document results from a suite of experiments chosen to provide a comparison of the performance of the \ac{mekf}, the Imperfect-\ac{iekf}, the \ac{tfgiekf}, an \ac{eqf} based on the $\grpD$ symmetry~\cite{Fornasier2022EquivariantBiases} termed TG-\ac{eqf}, an \ac{eqf} based on the $\grpA$ symmetry termed DP-\ac{eqf}, and an \ac{eqf} based on the $\grpB$ symmetry termed SD-\ac{eqf}. \deleted{, and finally an implementation of the \ac{tfgiekf} in~\cite{Barrau2022TheProblems} (implemented according to the author's original manuscript, and verified to behave the same as an \ac{eqf} based on the $\grpC$ symmetry)} \added{Moreover, given the global nature of the measurement, we included in the comparison a left-sided Imperfect-\ac{iekf} which directly uses the measurement in \equref{confout}. Note that each of the aforementioned \acp{eqf} have been implemented including the reset step discussed in \secref{prelim}, whereas the Imperfect-\acp{iekf} and the \ac{tfgiekf} have been implemented without reset step, according to their original implementation, discussed in~\cite{7523335, Barrau2022TheProblems} respectively}.
We undertake two separate experimental analyses. 
In the first experiment, we undertake a Monte-Carlo simulation of an \ac{uav} equipped with an \ac{imu} receiving acceleration and angular velocity measurements at $200$\si[per-mode = symbol]{\hertz} and receiving global position measurements at $10$\si[per-mode = symbol]{\hertz}, simulating a \ac{gnss} receiver. 
In the second experiment, we compare all the filters with real data from the INSANE dataset~\cite{Brommer2024TheScenarios}.

\subsection{UAV Flight Simulation}



\begin{figure*}
    \centering
    \subfloat[Average \acs{rmse} and sample variance (shaded) of the filters' states, and full filter energy.]{%
      \includegraphics[width=\linewidth]{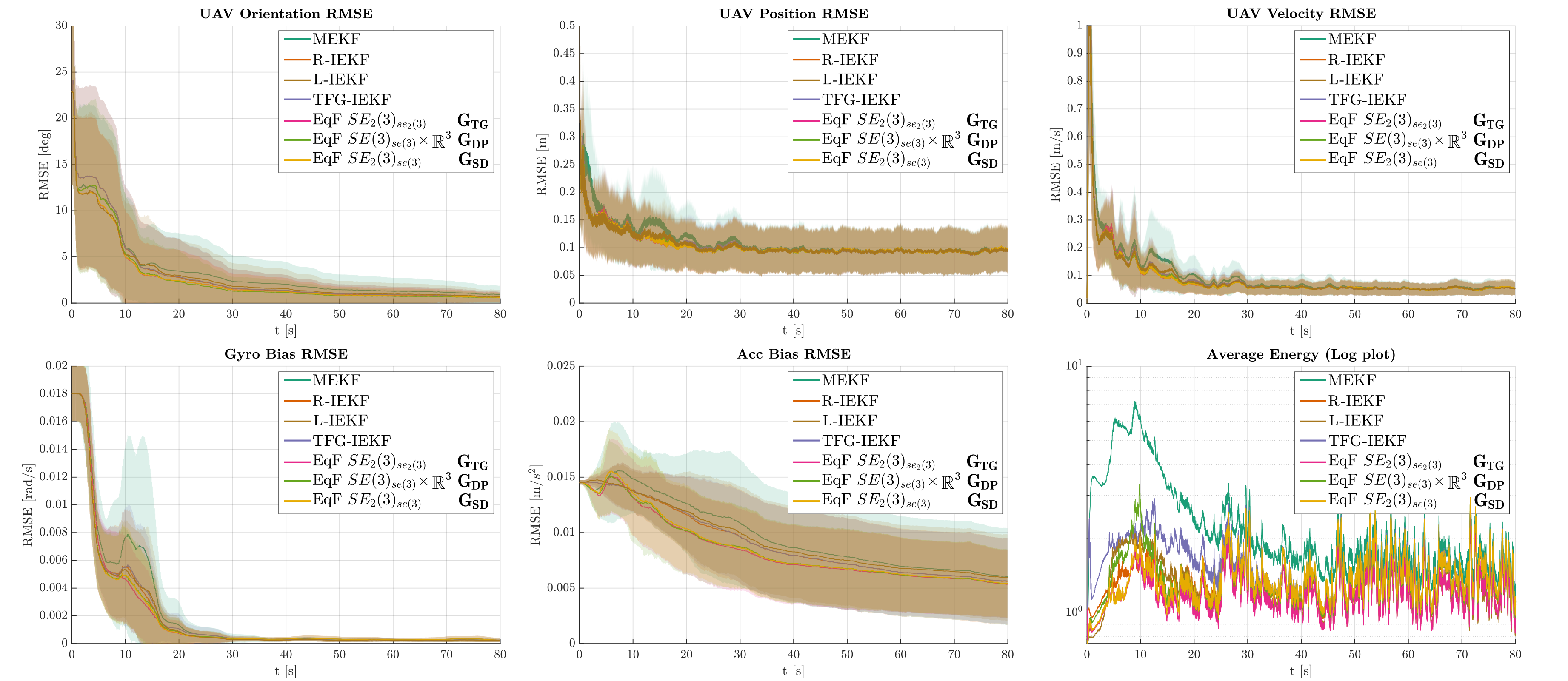}\label{fig:mc_errors}
    }\\
    \subfloat[Orientation and position states evolution, and innovation energy for the real-world \ac{uav} flight experiment.]{%
      \includegraphics[width=\linewidth]{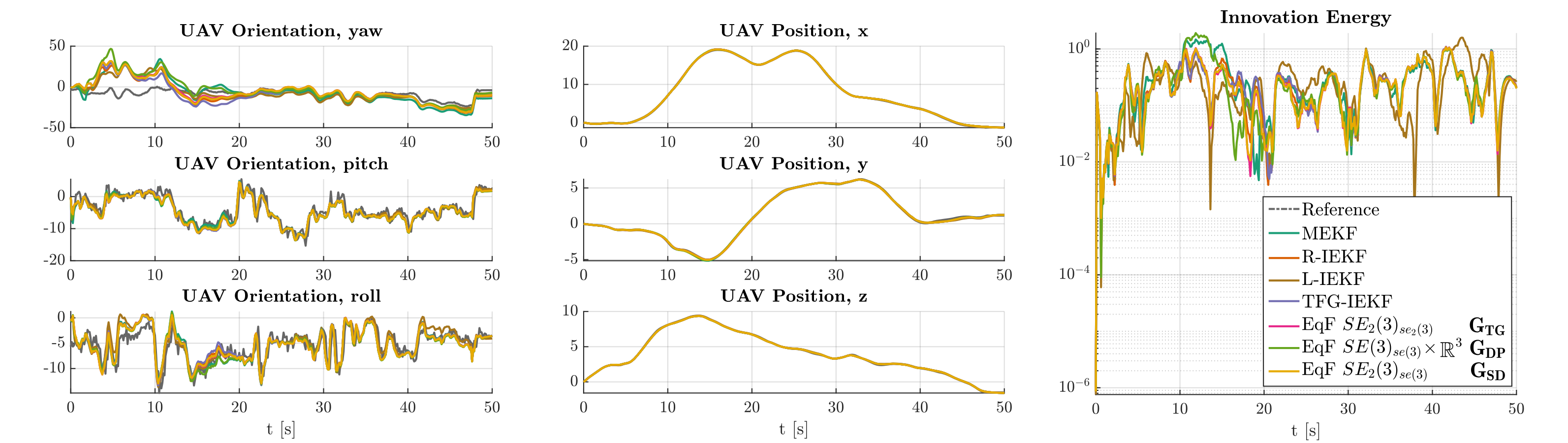}\label{fig:insane_errors}
    }
    \caption{Simulation and real-world experiments' results. {\color{DarkOrange2} Orange: R-\ac{iekf}}. {\color{Chocolate4} Brown: L-\ac{iekf}}. {\color{SlateBlue4} Purple: \ac{tfgiekf}}. {\color{Magenta2} Magenta: TG-\ac{eqf}}. {\color{OliveDrab3} Green: DP-\ac{eqf}}. {\color{Goldenrod1} Yellow: SD-\ac{eqf}}.}
\end{figure*}

In this experiment, we conducted a Monte-Carlo simulation, including four hundred runs of a simulated \ac{uav} equipped with an \ac{imu} and receiving global position measurements simulating a \ac{gnss} receiver. In order to simulate realistic flight conditions, we selected the initial $80$\si[per-mode = symbol]{\second} from four sequences in the Euroc dataset's vicon room~\cite{Burri2016TheDatasets} as reference trajectories. For each sequence, we generated a hundred runs, incorporating synthetic IMU data  generated interpolating and differentiating the Vicon poses, and position measurements while varying the initial conditions for the position (distributed normally around zero with $1$\si[per-mode = symbol]{\meter} standard deviation per axis) and the attitude (distributed normally around zero with $20$\si[per-mode = symbol]{\degree} standard deviation per axis). This experiment considers common realistic condition. For application of the semi-direct product symmetries in scenarios of difficult alignment we refer the reader to the authors' related work~\cite{Fornasier2022EquivariantBiases, Fornasier2022OvercomingCalibration, Scheiber2023RevisitingApproach, Fornasier2023MSCEqF:Navigation}.

The ground truth \ac{imu} biases are randomly generated every run following a Gaussian distribution with standard deviation of $0.01$\si[per-mode = symbol]{\radian\per\second\sqrt{\second}} for the gyro bias and $0.01$\si[per-mode = symbol]{\meter\per\second\squared\sqrt{\second}} for the accelerometer bias. 
To simulate realistic global position measurement, additive Gaussian noise with a standard deviation of $0.2$\si[per-mode = symbol]{\meter} per axis is added.

For a fair comparison, we were careful to use the same prior distributions and noise parameters for all filters. 
This includes accounting for the different scaling and transformations of noise due to the input and state parametrizations for the different geometries. 
Similarly, each filter shares the same input and output measurement noise covariance adapted to the particular geometry of the filter. 
The validity of the noise models can be verified in the Average Energy plot (\figref{mc_errors}), which plots the \acf{anees} \cite{Li2012EvaluationTests}. 
Here, all filters initialize with unity \ac{anees}, demonstrating that the prior sampling and observer response corresponds to the stochastic prior used, and all filters converge towards unity \ac{anees} as expected from a filter driven by Gaussian noise. 
All the filters are initialized at the identity (zero attitude, zero position, zero velocity, and zero biases).

The primary plots in \figref{mc_errors} show the \acs{rmse} values for the navigation states (on the top) and the bias states (on the bottom). 
It is clear that the \ac{mekf} filter demonstrates worse performance than the modern geometric filters. 
There is little difference visible in the transient and asymptotic error response of the navigation states for the modern filters. 
The remaining attitude error is due to yaw error.
The position and velocity errors converge to the noise limits of the measurement signals. 
In contrast, there are clear differences visible in the transient response of the bias states. 
To make the difference in the transient response clearer,~\tabref{transient_values}, shows the average \acs{rmse} values of each filter for the first $30$\si[per-mode = symbol]{\second} of the trajectory and the percentage improvement of the geometric filters with respect to the \acs{mekf}. Additionally,~\tabref{transient_times} shows the time taken by each filter to reach an average \acs{rmse} below $10\%$ of the maximum \acs{rmse} value. Based on these results, the filters split roughly into three categories: 
the three filters with semi-direct bias symmetry (TG-\ac{eqf}, DP-\ac{eqf} and SD-\ac{eqf}) which appear to display the best transient response in any state. 
The \acp{iekf} and \ac{tfgiekf}, which have the $\SE_2(3)$ symmetry but do not use a semi-direct geometry for the bias geometry, have \deleted{almost identical} \added{similar} bias transient. 
The accelerometer bias, in particular, is clearly separated from the filters with the semi-direct group symmetry. 
Finally, the \ac{mekf} which suffers from not modeling the $\SE_2(3)$ symmetry at all. 

\begin{table*}
    \setlength{\tabcolsep}{4.5pt}
    \centering
    \small
    \caption{RMSE values and percentage of improvement of the geometric filters with respect to the MEKF for the first $30\text{s}$ of the UAV simulation, corresponding to the transient phase (T). Best values are in \textbf{bold}, second best values are \underline{underlined}. \added{In the table, R-\ac{iekf} and L-\ac{iekf} represent respectively the right-sided and the left-sided Imperfect-\ac{iekf}.}}
    \begin{tabular}{c c c c c c c c}
         \toprule
         \acs{rmse} (T) & \ac{mekf} & R-\ac{iekf} & \added{L-\ac{iekf}} & \ac{tfgiekf} & TG-\ac{eqf} & DP-\ac{eqf} & SD-\ac{eqf}\\
         \midrule
         {Orientation} & $0.1108\;(100\%)$ & $0.1076\;(97\%)$ &  \added{$0.1011\;(91\%)$} & $0.1082\;(98\%)$ & $\mathbf{0.0938\;(85\%)}$ & $0.0968\;(87\%)$ & $\underline{0.0939\;(85\%)}$ \\
         {Position} & $0.1433\;(100\%)$ & $0.1276\;(89\%)$ &  \added{$0.129\;(90\%)$} & $0.1290\;(90\%)$ & $\underline{0.1267\;(88\%)}$ & $0.1270\;(89\%)$ & $\mathbf{0.1263\;(88\%)}$ \\
         {Velocity} & $0.1728\;(100\%)$ & $0.1459\;(84\%)$ &  \added{$0.146\;(84\%)$} & $0.1488\;(86\%)$ & $\underline{0.1411\;(82\%)}$ & $0.1411\;(82\%)$ & $\mathbf{0.1399\;(81\%)}$ \\
         {Gyro bias} & $0.0054\;(100\%)$ & $0.0044\;(83\%)$ &  \added{$0.0045\;(83\%)$} & $0.0046\;(85\%)$ & $\underline{0.0043\;(81\%)}$ & $0.0043\;(81\%)$ & $\mathbf{0.0043\;(80\%)}$  \\
         {Acc bias} & $0.0134\;(100\%)$ & $0.0125\;(94\%)$ &  \added{$0.013\;(95\%)$} & $0.0126\;(94\%)$ & $\mathbf{0.0116\;(87\%)}$ & $\underline{0.0117\;(88\%)}$ & $0.0119\;(89\%)$  \\
         \bottomrule
    \end{tabular}
    \label{tab:transient_values}
\end{table*}
\begin{table*}
    \setlength{\tabcolsep}{3pt}
    \centering
    \small
    \caption{Transient times taken by each filter to reach an average RMSE below $10\%$ of the maximum RMSE value. The accelerometer bias is excluded since the average RMSE never reached a value below $10\%$ of the maximum RMSE value. \added{Best values are in \textbf{bold}, second best values are \underline{underlined}. In the table, R-\ac{iekf} and L-\ac{iekf} represent respectively the right-sided and the left-sided Imperfect-\ac{iekf}.}}
    \begin{tabular}{c c c c c c c c}
         \toprule
         Transient time & \ac{mekf} & R-\ac{iekf} & \added{L-\ac{iekf}} & \ac{tfgiekf} & TG-\ac{eqf} & DP-\ac{eqf} & SD-\ac{eqf}\\
         \midrule
         {Orientation} & $32.090\;(100\%)$ & $23.095\;(72\%)$ & \added{$26.595\;(83\%)$} & $23.995\;(75\%)$ & $\underline{21.295\;(66\%)}$ & $\mathbf{20.395\;(64\%)}$ & $  \underline{21.295\;(66\%)}$\\
         {Position} & $5.095\;(100\%)$ & $\underline{4.795\;(94\%)}$ & \added{$\mathbf{4.595\;(90\%)}$} & $\underline{4.795\;(94\%)}$ & $4.995\;(98\%)$ & $\underline{4.795\;(94\%)}$ & $\underline{4.795\;(94\%)}$\\
         {Velocity} & $16.295\;(100\%)$ & $\mathbf{12.495\;(77\%)}$ & \added{$15.895\;(98\%)$} & $\mathbf{12.495\;(77\%)}$ & $\mathbf{12.495\;(77\%)}$ & $\underline{12.695\;(78\%)}$ & $\mathbf{12.495\;(77\%)}$\\
         {Gyro bias} & $17.095\;(100\%)$ & $\underline{16.095\;(94\%)}$ & \added{$16.395\;(96\%)$} & ${16.295\;(95\%)}$ & $\mathbf{15.895\;(93\%)}$ & ${16.195\;(95\%)}$ & $\mathbf{15.895\;(93\%)}$\\
         \bottomrule
    \end{tabular}
    \label{tab:transient_times}
\end{table*}
\begin{table*}
    \setlength{\tabcolsep}{12.5pt}
    \centering
    \caption{\acs{anees} for the first $30$\si[per-mode = symbol]{\second}, corresponding to the transient phase (T) and for the remaining of the trajectory length, corresponding to the asymptotic phase (A). Best values are in \textbf{bold}\added{, second best values are \underline{underlined}. In the table, R-\ac{iekf} and L-\ac{iekf} represent respectively the right-sided and the left-sided Imperfect-\ac{iekf}.}}
    \begin{tabular}{c c c c c c c c}
    \toprule
    \acs{anees} & \ac{mekf} & R-\ac{iekf} & \added{L-\ac{iekf}} & \ac{tfgiekf} & TG-\ac{eqf} & DP-\ac{eqf} & SD-\ac{eqf} \\
    \midrule
    Transient (T) & $3.11$ & $1.36$ & \added{$1.39$} & $1.71$ & $\mathbf{1.20}$ & $1.44$ & $\underline{1.32}$ \\
    Asymptotic (A) & $1.69$ & $1.40$ & \added{$\mathbf{1.20}$} & $1.43$ & $\underline{1.22}$ & $1.42$ &$1.44$ \\
    \bottomrule
    \end{tabular}
    \label{tab:anees}
\end{table*}

The average energy plot provides an additional important analysis tool. 
This plot shows the \ac{anees}~\cite{Li2012EvaluationTests} defined as 
\[
{\text{ANEES} = \frac{1}{nM}\sum_{i = 1}^{M}\Vector{}{\varepsilon}{i}^T\mathbf{\Sigma}_{i}^{-1}\Vector{}{\varepsilon}{i}},
\]
where $\Vector{}{\varepsilon}{}$ is the specific filter error state, $\mathbf{\Sigma}$ is the error covariance, ${M = 400}$ is the number of runs in the Monte-Carlo simulation, and $n$ is the dimension of the state space.  
The \ac{anees} provides a measure of the consistency of the filter estimate. 
An \ac{anees} of unity means that the observed error variance corresponds exactly to the estimated covariance of the information state. 
When \ac{anees} is larger than unity, it indicates that the filter is overconfident; that is, the observed error is larger than the estimate of the state covariance predicts. 
All `pure' extended Kalman filters tend to be overconfident since their derivation ignores linearization errors in the model. 
The closer to an \ac{anees} of unity that a filter manages is directly correlated to the consistency of the filter estimate and is usually linked to smaller linearization errors. 
To provide numeric results, we have averaged the \ac{anees} values over the transient and asymptotic sections of the filter response and presented them in~\tabref{anees}. 
Here, it is clear that the TG-\ac{eqf} is superior, \added{with a consistent \ac{anees} for the whole duration of the experiment,} the \deleted{four} \added{five other} filters \added{R-}\ac{iekf}, \added{L-\ac{iekf}} \ac{tfgiekf}, DP-\ac{eqf} and SD-\ac{eqf} are similar \added{in the transient phase, with the L-\ac{iekf} that improves in the asymptotic phase.} \deleted{, and} \added{Finally,} the \ac{mekf} is worst. 
The \ac{anees} of the \ac{mekf} diverges to over seven before converging, corresponding to an overconfidence of a factor of seven standard deviations in the state error. 
Such a level of overconfidence is dangerous in a real-world scenario and may indeed lead to divergence of the filter estimate in certain situations. 
Note that in practice, overconfidence of a filter is avoided by inflating the process noise model covariance to account for linearization error in the model. 
A more consistent filter requires a smaller covariance inflation and has correspondingly more confidence in its model than a filter that is less consistent. 

In conclusion, the TG-\ac{eqf} exhibits the best convergence rate, particularly in orientation and \ac{imu} biases, as well as the best consistency of all the filters. Hence, the TG-\ac{eqf} is the filter we recommend using for \ac{ins} problems. 
We believe that this performance can be traced back to the coupling of the \ac{imu} bias with the navigation states that are inherent in the semi-direct product structure of the symmetry group $\grpD$ and the exact linearization of the navigation error dynamics (\tabref{symmetries_overview}). 
Note that the bias states are poorly observable states and possess slow dynamics. 
Consequently, moving linearization error into these states heuristically appears better than leaving the linearization error in the main navigation states that are much more dynamic.

\subsection{Real-world UAV flight}

In this experiment, we compared the performance of the discussed filters in a real-world \ac{uav} flight scenario from the INSANE dataset~\cite{Brommer2024TheScenarios}. In particular, in this experiment, a quadcopter is flying for $50$\si[per-mode = symbol]{\meter} at a maximum height of $13$\si[per-mode = symbol]{\meter} covering an area of roughly $200$\si[per-mode = symbol]{\meter\squared}, at a maximum speed of $3$\si[per-mode = symbol]{\meter\per\second}. The \ac{uav} is receiving \ac{imu} measurements at $200$\si[per-mode = symbol]{\hertz}, as well as measurements from and RTK-\ac{gnss} receiver at $8$\si[per-mode = symbol]{\hertz} with an accuracy between $0.1$\si[per-mode = symbol]{\meter} and $0.6$\si[per-mode = symbol]{\meter}. The position and orientation reference has been obtained as described in~\cite{Brommer2024TheScenarios} from raw sensor measurements of two RTK-\ac{gnss} and a magnetometer.

Similar to the previous experiments, all the filters share the same tuning parameters.

\figref{insane_errors} shows the evolution of each filter orientation and position estimates, as well as the innovation energy, commonly referred to as the Normalised Innovation Squared (NIS) error 
\[
{\text{NIS} = \frac{1}{n}\Vector{}{r}{}^T\mathbf{S}^{-1}\Vector{}{r}{}},
\]
where ${\Vector{}{r}{}}$ is the specific measurement residual of dimension $n$ computed via the output action $\rho$ and $\mathbf{S}$ is the innovation covariance. 
The results in \figref{insane_errors} show that the high level conclusions from the simulations are confirmed on real data. 
There are slight differences between filters in the plotted results but due to the lack of accurate ground truth all that can be deduced is that all the filters provide high quality real-world INS solutions. 
This is not surprising since the \ac{mekf} is the industry standard and is know to perform well in practice and the more modern filters are expected to improve on this performance. 

%% file: sections/conclusion.tex
\section{Conclusion}\label{sec:conc}

This study investigates \acl{ins} filter design from the perspective of symmetry. 
We establish a unifying framework, demonstrating that various modern \ac{ins} filter variants can be interpreted as \aclp{eqf} applied to distinct choices of symmetry, with the group structure being the only difference among those filter variants.
With specific application to position measurements, we demonstrated that fixed-frame measurements can be reformulated as body-frame relative measurements. 
This allows one to exploit the equivariance of the output, ensuring third-order linearization error in the measurement equations. 

We discussed and presented different symmetry groups acting on the state-space of the \ac{ins} problem.
Novel symmetries are introduced alongside analysis of similarities and differences in the context of filter design. Furthermore, we showed how different choices of symmetries lead to filters with different linearized error dynamics, and how the $\grpD$ symmetry yields exact linearization of the navigation error, shifting all the lineraisation error into the bias state dynamics. 

Comparative performance studies in simulation, and real-world of a vehicle equipped with an \ac{imu} and receiving position measurement from a \ac{gnss} receiver highlighted that any of the \added{R-}\ac{iekf}, \added{L-\ac{iekf}}, \ac{tfgiekf}, TG-\ac{eqf}, DP-\ac{eqf}, and SD-\ac{eqf} are good candidates for high performance \ac{ins} filter design with the TG-\ac{eqf} demonstrating superior performance.

\section{Acknowledgment}

\added{The authors would like to thank Martin Scheiber for his invaluable support and insightful discussions, which helped achieve the results presented in the manuscript. The authors would also like to thank the reviewers for their insightful critique that has substantially improved the contribution of the paper.}

%% file: sections/appendix.tex
\section{Mathematical Preliminaries}
In this section, we provide a quick overview of fundamental concepts in Lie theory.

\subsection{Lie theory}
A Lie group $\grpG$ is a smooth manifold endowed with a smooth group structure.
For any $X, Y \in \grpG$, the group multiplication is denoted $XY$, the group inverse $X^{-1}$ and the identity element $I$. 

Given a Lie group $\grpG$, $\calG$ denotes the $\grpG$-Torsor\cit{Mahony2013ObserversSymmetry}, which is defined as the set of elements of $\grpG$ (the underlying manifold), but without the group structure.

For a given Lie group $\grpG$, the Lie algebra $\gothg$ can be modelled as a vector space corresponding to the tangent space at the identity of the group, together with a bilinear non-associative map ${[\cdot, \cdot] \AtoB{\gothg}{\gothg}}$ called the \emph{Lie bracket}.
For a matrix Lie group, the Lie bracket is equal to the matrix commutator:
\begin{equation*}
    \left[\eta, \kappa \right] = \eta \kappa - \kappa \eta ,
\end{equation*}
for any $\eta, \kappa \in \gothg \subset \R^{n \times n}$.

The Lie algebra $\gothg$ is isomorphic to a vector space $\R^{n}$ of dimension ${n =\mathrm{dim}\left(\gothg\right)}$.
Define the \emph{wedge} map and its inverse, the \emph{vee} map, as linear mappings between the vector space and the Lie algebra:
\begin{equation*}
    \left(\cdot\right)^{\wedge} \AtoB{\R^{n}}{\gothg} ,\qquad \left(\cdot\right)^{\vee} \AtoB{\gothg}{\R^{n}} . 
\end{equation*} 

For any $X, Y \in \grpG$, define the left and right translation
\begin{align*}
    &\textrm{L}_{X} \AtoB{\grpG}{\grpG}, \qquad \textrm{L}_{X}\left(Y\right) = XY ,\\
    &\textrm{R}_{X} \AtoB{\grpG}{\grpG}, \qquad \textrm{R}_{X}\left(Y\right) = YX . 
\end{align*}

The Adjoint map for the group $\grpG$, $\Ad_X \AtoB{\gothg}{\gothg}$ is defined by
\begin{equation*}
    \Adsym{X}{\bm{u}^{\wedge}} = \td \textrm{L}_{X} \td \textrm{R}_{X^{-1}}\left[\bm{u}^{\wedge}\right] ,
\end{equation*}
for every $X \in \grpG$ and ${\bm{u}^{\wedge} \in \gothg}$, where $\td \textrm{L}_{X}$, and $\td \textrm{R}_{X}$ denote the differentials of the left, and right translation, respectively.
Given particular wedge and vee maps for a matrix Lie group, the Adjoint matrix is defined as the map ${\AdMsym{X} \AtoB{\R^{n}}{\R^{n}}}$
\begin{equation*}
    \AdMsym{X}\bm{u} = \left(\Adsym{X}{\bm{u}^{\wedge}}\right)^{\vee} .
\end{equation*}

In addition to the Adjoint map for the group $\grpG$, the adjoint map for the Lie algebra $\gothg$ can be defined as the differential at the identity of the Adjoint map for the group $\grpG$.
The adjoint map for the Lie algebra $\ad_{\bm{u}^\wedge} \AtoB{\gothg}{\gothg}$ is given by 
\begin{equation*}
    \adsym{\bm{u}^\wedge}{\bm{v}^{\wedge}} = \left[\bm{u}^{\wedge}, \bm{v}^{\wedge}\right] ,
\end{equation*}
and is equivalent to the Lie bracket. 
Given particular wedge and vee maps for a matrix Lie group, we define the adjoint matrix $\adMsym{\bm{u}} \AtoB{\R^{n}}{\R^{n}}$ to be
\begin{equation*}
    \adMsym{\bm{u}}\bm{v} = \left(\bm{u}^{\wedge}\bm{v}^{\wedge} - \bm{v}^{\wedge}\bm{u}^{\wedge}\right)^{\vee} = \left[\bm{u}^{\wedge}, \bm{v}^{\wedge}\right]^{\vee} .
\end{equation*}

For all $\Vector{}{u}{}, \Vector{}{v}{} \in \R^n$ and $X \in \grpG$, the two adjoints commute:
\begin{align*}
    \Adsym{X}{\adsym{\Vector{}{u}{}^{\wedge}}{\Vector{}{v}{}^{\wedge}}} &= \adsym{\left(\Adsym{X}{\Vector{}{u}{}^{\wedge}}\right)}{\Adsym{X}{\Vector{}{v}{}^{\wedge}}},\\
    \AdMsym{X}\adMsym{\Vector{}{u}{}}\Vector{}{v}{} &= \adMsym{\AdMsym{X}{\Vector{}{u}{}}}\AdMsym{X}\Vector{}{v}{} ,
\end{align*}

\subsection{Semi-direct product groups}
A semi-direct product group $\mathbf{G} \ltimes \mathbf{H}$ can be seen as a generalization of the direct product group $\grpG \times \grpH$ where the underlying set is given by the cartesian product of two groups $\grpG$ and $\grpH$. 
Contrary to the direct product, in the semi-direct product, the group multiplication is defined with the group $\grpG$ that acts on a group $\grpH$ by a group automorphism. 
Note that the semi-direct product group $\mathbf{G} \ltimes \mathbf{H}$ with the trivial automorphism corresponds to the direct product group $\grpG \times \grpH$.

In this work we will consider a semi-direct product symmetry group \cite{ng2019attitude}, \cite{Ng2020EquivariantGroups}, \cite{Ng2020PoseKinematics} $\sdpgrpG := \mathbf{G} \ltimes \gothg$ where $\gothg$ is the Lie algebra of $\grpG$, or a subalgebra of $\grpG$, that is, a vector space group under addition. 
Let $A,B \in \grpG$ and $a,b \in \gothg$ and define $X = \left(A, a\right)$ and $Y = \left(B, b\right)$ elements of the symmetry group $\sdpgrpG$. 
Group multiplication is defined to be the semi-direct product ${YX = \left(BA, b + \Adsym{B}{a}\right)}$. 
The inverse element is ${X^{-1} = \left(A^{-1}, -\Adsym{A^{-1}}{a}\right)}$ and  identity element is ${\left(I, 0\right)}$.

\subsection{Lie group action and homogeneous space}
A right group action of a Lie group $\grpG$ on a differentiable manifold $\calM$ is a smooth map ${\phi \AtoB{\grpG\times \calM}{\calM}}$ that satisfies 
\begin{equation*}
    \phi\left(I, \xi\right) = \xi ,\qquad
    \phi\left(X, \phi\left(Y, \xi\right)\right) = \phi\left(YX, \xi\right) ,
\end{equation*} 
for any $X,Y \in \grpG$ and $\xi \in \calM$.
A right group action $\phi$ induces a family of diffeomorphism ${\phi_X \AtoB{\calM}{\calM}}$ and smooth projections ${\phi_{\xi} \AtoB{\grpG}{\calM}}$.
The group action $\phi$ is said to be transitive if the induced projection $\phi_{\xi}$ is surjective.
In this case, $\calM$ is a homogeneous space of $\grpG$.

\subsection{Important Lie groups}

The \emph{special orthogonal group} $\SO(3)$ is the Lie group of 3D rotations in space. The special orthogonal group and its Lie algebra are defined as follows:
\begin{align*}
    \SO(3) &= \set{\mathbf{A} \in \R^{3\times3}}{
    \mathbf{A} \mathbf{A}^\top = \mathbf{I}_3, \; \det(\mathbf{A} ) = 1}, \\
    \so(3) &= \set{\Vector{}{\omega}{}^{\wedge} \in \R^{3\times3}}{
    \Vector{}{\omega}{}^{\wedge} = - {\Vector{}{\omega}{}^{\wedge}}^\top, \; \Vector{}{\omega}{}^{\wedge}\Vector{}{v}{} = \Vector{}{\omega}{} \times \Vector{}{v}{}},
\end{align*}
where $\times$ represents the cross-product. 

The \emph{special Euclidean group} $\SE(3)$ is the group of 3D rigid body transformation in space. Note that $\SE(3)$ is defined as a semi-direct product group, specifically, $\SE(3) \coloneq \SO(3) \ltimes \R^6$. The special Euclidean group and its Lie algebra are defined as follows:
\begin{align*}
    \SE(3) &= \set{
    \begin{bmatrix}
        \mathbf{A} & \Vector{}{b}{} \\ \mathbf{0}_{1\times 3} & 1
    \end{bmatrix}
    \in \R^{4\times4}}{
    \mathbf{A} \in \SO(3), \; \Vector{}{b}{} \in \R^3}, \\
    \se(3) &= \set{
    \begin{bmatrix}
        \Vector{}{\omega}{}^{\wedge} & \Vector{}{w}{} \\ \mathbf{0}_{1\times 3} & 0
    \end{bmatrix}
    \in \R^{4\times4}}{
    \Vector{}{\omega}{}^{\wedge} \in \so(3), \; \Vector{}{w}{} \in \R^3},
\end{align*}

The \emph{extended special Euclidean group} $\SE_2(3)$~\cite{barrau:tel-01247723, 7523335, Barrau2020APreintegration, Brossard2021AssociatingEarth} is an extension of the special Euclidean group. The group and its Lie algebra are defined as follows:
\begin{align*}
    \SE_2(3) &= \set{
    \begin{bmatrix}
        \mathbf{A} & \Vector{}{a}{} & \Vector{}{b}{} \\ 
        \mathbf{0}_{1\times 3} & 1 & 0\\
        \mathbf{0}_{1\times 3} & 0 & 1
    \end{bmatrix}
    \in \R^{5\times5}}{
    \mathbf{A} \in \SO(3), \; \Vector{}{a}{}, \Vector{}{b}{} \in \R^{3}}, \\
    \se_2(3) &= \set{ 
    \begin{bmatrix}
        \Vector{}{\omega}{}^{\wedge} & \Vector{}{v}{} & \Vector{}{w}{} \\ \mathbf{0}_{1\times 3} & 0 & 0\\
        \mathbf{0}_{1\times 3} & 0 & 0
    \end{bmatrix}
    \in \R^{5\times5}}{
    \Vector{}{\omega}{}^{\wedge} \in \so(3), \; \Vector{}{v}{}, \Vector{}{w}{} \in \R^{3}}.
\end{align*}

The \emph{homogenous Galilean group} $\HG(3)$ is the group of 3D rotation and relative velocity transformations. $\HG(3)$ is isomorphic with $\SE(3)$ but acts on physical velocities rather than physical translations. The homogeneous Galilean group and its Lie algebra are defined as follows:
\begin{align*}
    \HG(3) &= \set{
    \begin{bmatrix}
        \mathbf{A} & \Vector{}{a}{} \\ \mathbf{0}_{1\times 3} & 1
    \end{bmatrix}
    \in \R^{4\times4}}{
    \mathbf{A} \in \SO(3), \; \Vector{}{a}{} \in \R^3}, \\
    \mathfrak{hg}(3) &= \set{
    \begin{bmatrix}
        \Vector{}{\omega}{}^{\wedge} & \Vector{}{v}{} \\ \mathbf{0}_{1\times 3} & 0
    \end{bmatrix}
    \in \R^{4\times4}}{
    \Vector{}{\omega}{}^{\wedge} \in \so(3), \; \Vector{}{v}{} \in \R^3},
\end{align*}

\section{Linearization Error and Equivariant Filter Design with Different Symmetries}\label{sec:append}

In this section, we explicitly derive the linearization error in the error dynamics and the related filter matrices of each filter presented in \tabref{symmetries_overview}.

\subsection{Logarithm map of semi-direct product group}
As mentioned in \secref{le_analysis}, for the semi-direct product groups $\grpG\ltimes\gothg$ where $\gothg$ is the Lie algebra of $\grpG$, we will use a matrix realization to compute the exponential algebraically.  
Define $X=(C,\gamma)\in\grpG\ltimes\gothg$, then $X$ has a matrix representation given by
\begin{align*}
    X = \begin{bNiceArray}{w{c}{0.55cm}w{c}{0.45cm}:w{c}{0.45cm}}[margin]
        \Block{2-2}{\AdMsym{C}} & & \Block{2-1}{\gamma^\vee}\\
        & & \Bstrut\\
        \hdottedline
        \Block{1-2}{\mathbf{0}} & & 1\Tstrut
    \end{bNiceArray}\in \R^{(\dim \gothg + 1)\times (\dim \gothg + 1)}
\end{align*}
The logarithm is then given by
\begin{align*}
    \log_{\grpG\ltimes\gothg}(X) = \begin{bNiceArray}{w{c}{0.75cm}w{c}{0.45cm}:w{c}{2.8cm}}[margin]
        \Block{2-2}{\adMsym{\log_\grpG(C)}} & & \Block{2-1}{J_l(\log_\grpG(C))^{-1}\gamma^\vee}\\
        & & \Bstrut\\
        \hdottedline
        \Block{1-2}{\mathbf{0}} & & 0\Tstrut
    \end{bNiceArray},
\end{align*}
where $J_l(\log_\grpG(C))$ is the left Jacobian of $\log_\grpG(C)$,
\begin{align*}
    J_l(\log_\grpG(C)) = \sum_{k=0}^{\infty}\frac{1}{(k+1)!}{\adMsym{\log_\grpG(C)}}^k.
\end{align*}

\subsection{\ac{mekf} linearization error and filter design}
\subsubsection{Overview}
The state space is defined to be ${\calM \coloneqq \mathcal{SO}(3)\times \R^3\times \R^3\times \R^3\times \R^3}$ with ${\xi \coloneqq \left(\Rot{}{},\Vector{}{v}{},\Vector{}{p}{},\Vector{}{b}{\bm{\omega}}, \Vector{}{b}{a}\right)\in\calM}$.
Choose the origin to be ${\mathring{\xi} \coloneqq \left(\eye_3, \Vector{}{0}{3\times1},\Vector{}{0}{3\times1},\Vector{}{0}{3\times1},\Vector{}{0}{3\times1}\right)\in\calM}$.
The velocity input is given by ${u \coloneqq \left(\Vector{}{\omega}{}, \Vector{}{a}{},\Vector{}{\tau}{\bm{\omega}},\Vector{}{\tau}{a}\right)}$.

The symmetry group of \ac{mekf} is given by ${\grpG_\mathbf{O} \coloneqq \SO(3)\times \R^{12}}$.
Define the filter state $\hat{X} = (\hat{A},\hat{a},\hat{b},\hat{\alpha},\hat{\beta})\in\grpG_\mathbf{O}$, where $\hat{A}\in\SO(3)$ and $\hat{a},\hat{b},\hat{\alpha},\hat{\beta}\in\R^3$.
The state estimate is given by
\begin{equation}
        \hat{\xi} \coloneqq \phi(\hat{X},\mathring{\xi}) = (\hat{A} ,\hat{a} ,\hat{b}, \hat{\alpha}, \hat{\beta}) = \left(\hatRot{}{},\hatVector{}{v}{},\hatVector{}{p}{},\hatVector{}{b}{\bm{\omega}}, \hatVector{}{b}{a}\right). 
\end{equation}
The state error is defined as 
\begin{equation}\label{eq:mekf_err}
    \begin{split}
        e \coloneqq \phi(\hat{X}^{-1},\xi) &= \left(\Rot{}{}\hat{A}^\top, \Vector{}{v}{} - \hat{a}, \Vector{}{p}{} - \hat{b}, \Vector{}{b}{\omega} - \hat{\alpha}, \Vector{}{b}{a} - \hat{\beta}\right) ,\\
        &= \left(\Rot{}{}\hatRot{}{}^\top, \Vector{}{v}{} - \hatVector{}{v}{}, \Vector{}{p}{} - \hatVector{}{p}{}, \Vector{}{b}{\omega} - \hatVector{}{b}{\omega}, \Vector{}{b}{a} - \hatVector{}{b}{a}\right).
    \end{split}
\end{equation}

\subsubsection{Error dynamics}
The error dynamics related to \equref{mekf_err} for each state is given by 
\begin{align*}
    \dot{e}_R &= \ddt (\Rot{}{} \hatRot{}{}^\top)\\
        &= \Rot{}{}(\Vector{}{\omega}{} - \Vector{}{b}{\bm{\omega}})^\wedge \hatRot{}{}^\top - \Rot{}{}\hatRot{}{}^\top\hatRot{}{}(\Vector{}{\omega}{} - \hatVector{}{b}{\bm{\omega}})^\wedge \hatRot{}{}^\top\\
        &=\Rot{}{}(\Vector{}{\omega}{} - \Vector{}{b}{\bm{\omega}}-\Vector{}{\omega}{}+\hatVector{}{b}{\bm{\omega}})^\wedge \hatRot{}{}^\top\\
        &=-e_R \hatRot{}{} e_{b_\omega}^\wedge \hatRot{}{}^\top\\
        &=-e_R \left(\hatRot{}{} e_{b_\omega}\right)^{\wedge};\\
    \dot{e}_v &= \ddt (\Vector{}{v}{} - \hatVector{}{v}{}) = \dot{\Vector{}{v}{}} - \dot{\hatVector{}{v}{}} \\
        &=\Rot{}{}(\Vector{}{a}{}-\Vector{}{b}{a})^\wedge + \Vector{}{g}{} - \hatRot{}{}(\Vector{}{a}{} - \hatVector{}{b}{a})^\wedge - \Vector{}{g}{}\\
        & = \Rot{}{}(\Vector{}{a}{}-\Vector{}{b}{a})^\wedge - \hatRot{}{}(\Vector{}{a}{} - \hatVector{}{b}{a})^\wedge\\
        & = e_R\hatRot{}{}(\Vector{}{a}{} - \Vector{}{b}{a}) - \hatRot{}{}(\Vector{}{a}{} -\hatVector{}{b}{a});\\
    \dot{e}_p &= \ddt (\Vector{}{p}{} - \hatVector{}{p}{}) = \dot{\Vector{}{p}{}} - \dot{\hatVector{}{p}{}}\\
        &=\Vector{}{v}{} - \hatVector{}{v}{} = e_v;\\
    \dot{e}_{b} &= \ddt(\Vector{}{b}{}-\hatVector{}{b}{}) = \mathbf{0}.
\end{align*}

The local coordinate chart ${\varepsilon = \log_{\grpG_\mathbf{O}} \circ\; \phi_{\mathring{\xi}}^{-1}(e)}$ for each state is given by
\begin{align*}
    &\varepsilon_R  \coloneqq  \log_{\SO(3)}(e_R)^\vee \in\R^3\\
    &\varepsilon_{v,p,{b_\omega},{b_a}} \coloneqq  e_{v,p,{b_\omega},{b_a}} \in\R^3.
\end{align*}

The linearization of $\dot{e}_R= \tD \exp(\varepsilon_R^\wedge)[\dot{\varepsilon}_R^{\wedge}]$ is given by 
\begin{align*}
    e_R\frac{\text{I}-\exp(-\ad_{\varepsilon_R^{\wedge}})}{\ad_{\varepsilon_R^{\wedge}}}\dot{\varepsilon}_R^{\wedge} &= -e_R \left(\hatRot{}{} \varepsilon_{b_\omega}\right)^{\wedge}\\
    (\text{I} + \mathcal{O}(\varepsilon_R^\wedge))\dot{\varepsilon}_R^{\wedge} &= \left(\hatRot{}{} \varepsilon_{b_\omega}\right)^{\wedge}\\
    \dot{\varepsilon}_R &= \hatRot{}{}\varepsilon_{b_\omega} + O({\varepsilon_R}^2).
\end{align*}

The linearization of $\dot{e}_v = \dot{\varepsilon}_v$ is given by 
\begin{align*}
    \dot{\varepsilon}_v & = e_R\hatRot{}{}(\Vector{}{a}{} - \Vector{}{b}{a}) - \hatRot{}{}(\Vector{}{a}{} -\hatVector{}{b}{a})\\
    &=(\text{I}+\varepsilon_R^\wedge + O({\varepsilon_R}^2))\hatRot{}{}(\Vector{}{a}{} - \Vector{}{b}{a}) - \hatRot{}{}(\Vector{}{a}{} - \hatVector{}{b}{a})\\
    &=\hatRot{}{}(\hatVector{}{b}{a} - \Vector{}{b}{a}) + \varepsilon_R^\wedge\hatRot{}{}(\Vector{}{a}{}-\Vector{}{b}{a}) + O({\varepsilon_R}^2)\\
    &=-\hatRot{}{}\varepsilon_{b_a} + \varepsilon_R^\wedge\hatRot{}{}(\Vector{}{a}{}-\varepsilon_{b_a}-\hatVector{}{b}{a}) + \mathcal{O}(\varepsilon^2)\\
    &=-\hatRot{}{}\varepsilon_{b_a} + \varepsilon_R^\wedge\hatRot{}{}(\Vector{}{a}{}-\hatVector{}{b}{a}) - \varepsilon_R^\wedge\hatRot{}{}\varepsilon_{b_a} + O({\varepsilon_R}^2)\\
    &= -(\hatRot{}{}(\Vector{}{a}{}-\hatVector{}{b}{a}))^\wedge\varepsilon_R -\hatRot{}{}\varepsilon_{b_a}  + \mathcal{O}(\varepsilon^2).
\end{align*}

The linearization of $\dot{e}_p = \dot{\varepsilon}_p$ is given by
\begin{align*}
    \dot{\varepsilon}_p = \varepsilon_v.
\end{align*}

The linearization of $\dot{e}_{b} = \dot{\varepsilon}_{b}$ is given by $\dot{\varepsilon}_{b}=\Vector{}{0}{}$.

\subsubsection{Filter design}
From the linearization error analysis, it is trivial to see that the linearized error state matrix ${\mathbf{A}_{t}^{0} \st \dotVector{}{\varepsilon}{} \simeq \mathbf{A}_{t}^{0}\Vector{}{\varepsilon}{}}$ is written
\begin{equation}
    \mathbf{A}_{t}^{0} = \begin{bNiceArray}{cw{c}{0.45cm}c:w{c}{1.15cm}w{c}{1.35cm}}[margin]
        \Block{3-3}{\prescript{}{1}{\mathbf{A}}} & & & -\hatRot{}{} & \Vector{}{0}{3 \times 3}\\
        & & & \Vector{}{0}{3 \times 3} & -\hatRot{}{}\\
        & & & \Vector{}{0}{3 \times 3} & \Vector{}{0}{3 \times 3}\Bstrut\\
        \hdottedline
        \Block{1-3}{\mathbf{0}_{6 \times 9}} & & & \Block{1-2}{\Vector{}{0}{6 \times 6}}\Tstrut
    \end{bNiceArray} \in \R^{15 \times 15}, \label{eq:At0_mekf}
\end{equation}
where
\begin{equation*}
    \prescript{}{1}{\mathbf{A}} = \begin{bmatrix}
    \Vector{}{0}{3 \times 3} & \Vector{}{0}{3 \times 3} & \Vector{}{0}{3 \times 3}\\
    -\left(\hatRot{}{}\left(\Vector{}{a}{} - \hatVector{}{b}{a}\right)\right)^{\wedge} & \Vector{}{0}{3 \times 3} & \Vector{}{0}{3 \times 3}\\
    \Vector{}{0}{3 \times 3} & \eye_3 & \Vector{}{0}{3 \times 3}\\
    \end{bmatrix} \in \R^{9 \times 9}.
\end{equation*}

Position measurements are linear with respect to the defined error; therefore, the output matrix ${\mathbf{C}^{0}}$ yields
\begin{equation}
    \mathbf{C}^{0} = \begin{bmatrix}
        \mathbf{0}_{3 \times 6} & \eye_3 & \mathbf{0}_{3 \times 6}\\
        \end{bmatrix} \in \R^{3 \times 15}. \label{eq:C0_mekf}
\end{equation}
It is straightforward to verify that the derived \acl{eqf} is equivalent to the well-known \ac{mekf}, and the \ac{eqf} state matrix in \equref{At0_mekf} corresponds directly to the state matrix of the \ac{mekf}~\cite[Sec. 7]{Sola2017QuaternionFilter}

\subsection{Imperfect-\ac{iekf}}
\subsubsection{Overview}
The state space is defined as ${\calM \coloneqq \mathcal{SE}_2(3)\times\R^6}$ with ${\xi \coloneqq (\Pose{}{},\Vector{}{b}{})\in\calM}$. 
One has ${\Pose{}{} = (\Rot{}{},\Vector{}{v}{},\Vector{}{p}{})\in\mathcal{SE}_2(3)}$ and ${\Vector{}{b}{}=(\Vector{}{b}{\bm{\omega}},\Vector{}{b}{a}) \in \R^6}$.
Choose the origin to be ${\mathring{\xi} = \left(\eye_5,\Vector{}{0}{6\times1}\right)\in\calM}$.
The velocity input is given by ${u \coloneqq \left(\Vector{}{\omega}{}, \Vector{}{a}{},\Vector{}{\tau}{\bm{\omega}},\Vector{}{\tau}{a}\right)}$.

The symmetry group of Imperfect-\ac{iekf} is given by ${\grpG_{\mathbf{ES}}:\SE_2(3)\times\R^6}$. Define the filter state ${\hat{X} = (\hat{C},\hat{\gamma})\in\grpG_{\mathbf{ES}}}$ with ${\hat{C} = (\hat{A},\hat{a},\hat{b})\in\SE_2(3)}$ and ${\hat{\gamma} = (\hat{\gamma_\omega}, \hat{\gamma_a})\in\R^6}$.
The state estimate is given by 
\begin{align}
    \hat{\xi} \coloneqq \phi(\hat{X},\mathring{\xi}) = (\hat{A}, \hat{\gamma}) = (\hatPose{}{}, \hatVector{}{b}{}). 
\end{align}
The state error is defined as 
\begin{equation}\label{eq:iekf_err}
\begin{split}
    e \coloneqq \phi(\hat{X}^{-1},\xi) &= (\Pose{}{}\hat{C}^{-1},\Vector{}{b}{}-\hat{\gamma}) ,\\
    &= (\Pose{}{}\hatPose{}{}^{-1},\Vector{}{b}{}-\hatVector{}{b}{}).
\end{split} 
\end{equation}

\subsubsection{Error dynamics}
The error dynamics related to \equref{iekf_err} for each state is given by 
\begin{align*}
    \dot{e}_R &= -e_R(\hatRot{}{}e_{b_\omega})^\wedge\; \text{\small{(Derivation same as \ac{mekf})}};\\
    \dot{e}_v &=\ddt(-\Rot{}{}\hatRot{}{}^\top + \Vector{}{v}{}) = \ddt(-\Rot{}{}\hatRot{}{}^\top\hatVector{}{v}{}+\Vector{}{v}{})\\
        &=-\dot{e}_R\hatVector{}{v}{} - e_R\dot{\hatVector{}{v}{}}+\dot{\Vector{}{v}{}}\\
        &=e_R(\hatRot{}{}e_{b_\omega})^\wedge \hatVector{}{v}{}-e_R \hatRot{}{}(\Vector{}{a}{}-\hatVector{}{b}{a}) -e_R \Vector{}{g}{} + \Rot{}{}(\Vector{}{a}{}-\Vector{}{b}{a})+\Vector{}{g}{}\\
        &=e_R(\hatRot{}{}e_{b_\omega})^\wedge \hatVector{}{v}{}-e_R\hatRot{}{}(\Vector{}{a}{}-\hatVector{}{b}{a}) -e_R \Vector{}{g}{}\\
        &\quad + e_R\hatRot{}{}(\Vector{}{a}{}-e_{b_a}+\hatVector{}{b}{a})+\Vector{}{g}{};\\
    \dot{e}_p &= \ddt (-\Rot{}{}\hatRot{}{}^\top\hatVector{}{p}{}+\Vector{}{p}{})\\
        &= -\dot{e}_R\hatVector{}{p}{} - e_R\dot{\hatVector{}{p}{}}+\dot{\Vector{}{p}{}}\\
        &= e_R(\hatRot{}{}e_{b_\omega})^\wedge \hatVector{}{p}{} - e_R\hatVector{}{v}{} + \Vector{}{v}{}\\
        &= e_R(\hatRot{}{}e_{b_\omega})^\wedge \hatVector{}{p}{} + e_v;\\
    \dot{e}_b &= \Vector{}{0}{}.
\end{align*}

The local coordinate chart ${\varepsilon = \log_{\grpG_\mathbf{ES}} \circ\; \phi_{\mathring{\xi}}^{-1}(e)}$ for each state is given by
\begin{align*}
    &\varepsilon_T  \coloneqq   \log_{\SE_2(3)}(\phi_{\mathring{\xi}}^{-1}(e_T))^\vee =  \log_{\SE_2(3)}(e_T)^\vee \in\R^9\\
    &\varepsilon_{b_\omega,b_a} \coloneqq  e_{b_\omega, b_a} \in\R^3.
\end{align*}

The linearization of $\dot{e}_R$ is the same as the derivation in \ac{mekf}, given by
\begin{align*}
    \dot{\varepsilon}_R &= \hatRot{}{}\varepsilon_{b_\omega} + O({\varepsilon_R}^2).
\end{align*}

The linearization of $\dot{e}_v = \dot{\varepsilon}_v+\mathcal{O}(\varepsilon^2)$ is given by 
\begin{align*}
    \dot{\varepsilon}_v &= -(\text{I}+\varepsilon_R^\wedge + O({\varepsilon_R}^2))(\hatRot{}{}\varepsilon_{b_\omega})^\wedge\hatVector{}{v}{}\\
    &\quad-(\text{I}+\varepsilon_R^\wedge + O({\varepsilon_R}^2))\hatRot{}{}(\Vector{}{a}{}-\hatVector{}{b}{a})\\
    &\quad -(\text{I}+\varepsilon_R^\wedge + O({\varepsilon_R}^2)) \Vector{}{g}{}\\
    &\quad+ (\text{I}+\varepsilon_R^\wedge + O({\varepsilon_R}^2))\hatRot{}{}(\Vector{}{a}{}-\varepsilon_{b_a}+\hatVector{}{b}{a})\\
    &\quad+\Vector{}{g}{}+\mathcal{O}(\varepsilon^2)\\
    &= -\hatVector{}{v}{}^\wedge \hatRot{}{} \varepsilon_{b_\omega} - \hatRot{}{}\varepsilon_{b_a}+\Vector{}{g}{}^\wedge\varepsilon_R + \mathcal{O}(\varepsilon^2).
\end{align*}

The linearization of $\dot{e}_p = \dot{\varepsilon}_p+\mathcal{O}(\varepsilon^2)$ is given by 
\begin{align*}
    \dot{\varepsilon}_p &= (\text{I}+\varepsilon_R^\wedge + O({\varepsilon_R}^2))(\hatRot{}{}e_{b_\omega})^\wedge\hatVector{}{p}{} + \varepsilon_v+\mathcal{O}(\varepsilon^2)\\
    &=\varepsilon_v - \hatVector{}{p}{}^\wedge\varepsilon_{b_\omega}+\mathcal{O}(\varepsilon^2).
\end{align*}

The linearization of $\dot{e}_{b} = \dot{\varepsilon}_{b}$ is given by $\dot{\varepsilon}_{b}=\Vector{}{0}{}$.

\subsubsection{Filter design}
The linearized error state matrix ${\mathbf{A}_{t}^{0} \st \dotVector{}{\varepsilon}{} \simeq \mathbf{A}_{t}^{0}\Vector{}{\varepsilon}{}}$ yields
\begin{equation}
    \mathbf{A}_{t}^{0} = \begin{bNiceArray}{cw{c}{0.45cm}c:w{c}{1.15cm}w{c}{1.35cm}}[margin]
        \Block{3-3}{\prescript{}{2}{\mathbf{A}}} & & & -\hatRot{}{} & \Vector{}{0}{3 \times 3}\\
        & & & -\hatVector{}{v}{}^{\wedge}\hatRot{}{} & -\hatRot{}{}\\
        & & & -\hatVector{}{p}{}^{\wedge}\hatRot{}{} & \Vector{}{0}{3 \times 3}\Bstrut\\
        \hdottedline
        \Block{1-3}{\mathbf{0}_{6 \times 9}} & & & \Block{1-2}{\Vector{}{0}{6 \times 6}}\Tstrut
    \end{bNiceArray} \in \R^{15 \times 15}, \label{eq:At0_iekf}
\end{equation}
where
\begin{equation}
    \prescript{}{2}{\mathbf{A}} = \begin{bmatrix}
    \Vector{}{0}{3 \times 3} & \Vector{}{0}{3 \times 3} & \Vector{}{0}{3 \times 3}\\
    \Vector{}{g}{}^{\wedge} & \Vector{}{0}{3 \times 3} & \Vector{}{0}{3 \times 3}\\
    \Vector{}{0}{3 \times 3} & \eye_3 & \Vector{}{0}{3 \times 3}\\
    \end{bmatrix} \in \R^{9 \times 9} \label{eq:2A}
\end{equation}

Position measurements formulated according to \equref{confout_vec} are equivariant, yielding the following output matrix
\begin{equation}
    \mathbf{C}^{\star} = \begin{bmatrix}
        \frac{1}{2}\left(y + \hatVector{}{p}{}\right)^{\wedge} & \mathbf{0}_{3 \times 3} & -\eye_3 & \mathbf{0}_{3 \times 6}\\
        \end{bmatrix} \in \R^{3 \times 15}. \label{eq:Cstar_iekf}
\end{equation}
It is trivial to verify that the state matrix in \equref{At0_iekf}, derived according to \acl{eqf} design principles, directly corresponds to the state matrix in the Imperfect-\ac{iekf}~\cite[Sec. 7]{doi:10.1177/0278364919894385}.

\subsection{TG-\ac{eqf}}
\subsubsection{Overview}
The state space is defined as ${\calM \coloneqq \mathcal{SE}_2(3)\times\R^9}$ with ${\xi \coloneqq (\Pose{}{},\Vector{}{b}{})\in\calM}$. 
One has ${\Pose{}{} = (\Rot{}{},\Vector{}{v}{},\Vector{}{p}{})\in\mathcal{SE}_2(3)}$ and ${\Vector{}{b}{}=(\Vector{}{b}{\bm{\omega}},\Vector{}{b}{a},\Vector{}{b}{\bm{\nu}}) \in \R^9}$.
Choose the origin to be ${\mathring{\xi} = \left(\eye_5,\Vector{}{0}{9\times1}\right)\in\calM}$.
The velocity input is given by ${u \coloneqq \left(\Vector{}{\omega}{}, \Vector{}{a}{}, \Vector{}{\nu}{}, \Vector{}{\tau}{\bm{\omega}}, \Vector{}{\tau}{a}, \Vector{}{\tau}{\nu}\right)}$.

The symmetry group of TG-\ac{eqf} is given by ${\grpG_{\mathbf{TG}}:\SE_2(3)\ltimes \se_2(3)}$. Define the filter state ${\hat{X} = (\hat{C},\hat{\gamma})\in\grpG_{\mathbf{TG}}}$ with ${\hat{C} = (\hat{A},\hat{a},\hat{b})\in\SE_2(3)}$ and ${\hat{\gamma} = (\hat{\gamma_\omega}, \hat{\gamma_a}, \hat{\gamma_\nu})^\wedge\in\se_2(3)}$.
The state estimate is given by 
\begin{align}
    \hat{\xi} \coloneqq \phi(\hat{X},\mathring{\xi}) = (\hat{C}, \Ad_{\hat{C}^{-1}}(-\hat{\gamma}^\vee)) = (\hatPose{}{}, \hatVector{}{b}{}). 
\end{align}
The state error is defined as 
\begin{align}\label{eq:tg_err}
e \coloneqq \phi(\hat{X}^{-1},\xi) &= (\Pose{}{}\hat{C}^{-1},\AdMsym{\hat{C}}(\Vector{}{b}{}+\Adsym{\hat{C}^{-1}}{\hat{\gamma}}^\vee))\\
&= (\Pose{}{}\hat{C}^{-1},\AdMsym{\hat{C}}\Vector{}{b}{} + \hat{\gamma}^{\vee})\\
&= (\Pose{}{}\hatPose{}{}^{-1},\AdMsym{\hatPose{}{}}(\Vector{}{b}{} - \hatVector{}{b}{}))
\end{align}

\subsubsection{Error dynamics}
\paragraph*{Navigation states}
The error dynamics related to \equref{tg_err} for the navigation states ${{e}_T = \Pose{}{}\hatPose{}{}^{-1}}$ is given by
\begin{align*}
    \dot{e}_T &= \dot{\Pose{}{}}\hatPose{}{}^{-1} - \Pose{}{}\hatPose{}{}^{-1}\dot{\hatPose{}{}}\hatPose{}{}^{-1}\\
    &=\Pose{}{}(\mathbf{W}-\mathbf{B}+\mathbf{N})\hatPose{}{}^{-1} + (\mathbf{G}-\mathbf{N})\Pose{}{}\hatPose{}{}^{-1} \\
    &\quad- e_T\hatPose{}{}(\mathbf{W}-\hat{\mathbf{B}}+\mathbf{N})\hatPose{}{}^{-1} - e_T(\mathbf{G}-\mathbf{N})\hatPose{}{}\hatPose{}{}^{-1}\\
    &=e_T\hatPose{}{}(\mathbf{W}-\mathbf{B}+\mathbf{N})\hatPose{}{}^{-1} - e_T\hatPose{}{}(\mathbf{W}-\hat{\mathbf{B}}+\mathbf{N})\hatPose{}{}^{-1}\\
    &\quad+(\mathbf{G}-\mathbf{N})e_T - e_T(\mathbf{G}-\mathbf{N})\\
    &= e_T\Adsym{\hatPose{}{}}{\hat{\mathbf{B}}-\mathbf{B}} + (\mathbf{G}-\mathbf{N})e_T - e_T(\mathbf{G}-\mathbf{N}).
\end{align*}
The above dynamics can be separate to two parts: ${\dot{e}_T = \dot{e}_{T_W} + \dot{e}_{T_G}}$, where ${\dot{e}_{T_W} = e_T\Adsym{\hatPose{}{}}{\hat{\mathbf{B}}-\mathbf{B}}}$ and ${\dot{e}_{T_G} = (\mathbf{G}-\mathbf{N})e_T - e_T(\mathbf{G}-\mathbf{N})}$.
The linearization can be derived separately for each part.

The local coordinate chart is given by ${\varepsilon = \log \circ \phi_{\mathring{\xi}}^{-1}(e)}$.
For each state, one has 
\begin{align*}
    e_T &= \exp_{\SE_2(3)}({\varepsilon_T}^\wedge),\\
    e_b &= \Adsym{{e_T}^{-1}}{(-J_l(\varepsilon_T)\varepsilon_b)^\wedge},
\end{align*}
where the exponential map $\exp_{\SE_2(3)\ltimes\se_2(3)}$ is derived from the semi-direct product structure, and $J_l(\varepsilon_T)$ is the left Jacobian of $\SE_2(3)$, given by 
\begin{align*}
    J_l(\varepsilon_T) = \sum_{k=0}^{\infty}\frac{1}{(k+1)!}\ad_{\varepsilon_T}^k.
\end{align*}

Recall that by definition \equref{tg_err} one has ${e_b^{\wedge} \coloneqq \Adsym{\hatPose{}{}}{(\Vector{}{b}{} - \hatVector{}{b}{})^{\wedge}}}$ with ${\Vector{}{b}{}^{\wedge} = \mathbf{B}}$, and ${\hatVector{}{b}{}^{\wedge} = \hat{\mathbf{B}}}$. Hence, for $\dot{e}_{T_W}$ one has 
\begin{align*}
    \dot{e}_{T_W} &= e_T\Adsym{\hatPose{}{}}{\hat{\mathbf{B}}-\mathbf{B}} = -e_Te_b^\wedge.
\end{align*}
Substituting the local coordinate yields
\begin{align}
    \tD\exp_{\SE_2(3)}({\varepsilon_T}^\wedge)[{\dot{\varepsilon}_{T_W}}^\wedge]&= -e_T \Adsym{{e_T}^{-1}}{(-J_l(\varepsilon_T)\varepsilon_b)^\wedge}\nonumber\\
    e_T \frac{\text{I}-\exp(-\ad_{{\varepsilon_T}^\wedge})}{\ad_{{\varepsilon_T}^\wedge}} {\dot{\varepsilon}_{T_W}}^\wedge &= -e_T \Adsym{{e_T}^{-1}}{(-J_l(\varepsilon_T)\varepsilon_b)^\wedge}\nonumber\\
    \Ad{e_T}\frac{\text{I}-\exp(-\ad_{{\varepsilon_T}^\wedge})}{\ad_{{\varepsilon_T}^\wedge}} {\dot{\varepsilon}_{T_W}}^\wedge &= (J_l(\varepsilon_T)\varepsilon_b)^\wedge. \label{eq:tg_temp1}
\end{align}
Because $\Ad_{e_T} = \Ad_{\exp({\varepsilon_T}^\wedge)}= \exp(\ad_{{\varepsilon_T}^\wedge})$, the term on the left side in \equref{tg_temp1} can be modified as 
\begin{align}
    \Ad_{e_T}\frac{\text{I}-\exp(-\ad_{{\varepsilon_T}^\wedge})}{\ad_{{\varepsilon_T}^\wedge}} 
    &= \exp(\ad_{{\varepsilon_T}^\wedge})\frac{\text{I}-\exp(-\ad_{{\varepsilon_T}^\wedge})}{\ad_{{\varepsilon_T}^\wedge}} \nonumber\\
    &= \frac{\exp(\ad_{{\varepsilon_T}^\wedge})-\text{I}}{\ad_{{\varepsilon_T}^\wedge}} \quad\text{(Expanding exp,)}\nonumber\\
    &= \sum_{k=0}^{\infty}\frac{1}{(k+1)!}\ad_{{\varepsilon_T}^\wedge}^k = J_l(\varepsilon_T) .\label{eq:jl}
\end{align}
Hence, for the linearization of $\dot{e}_{T_W}$, one has
\begin{align*}
    \dot{\varepsilon}_{T_W} = \varepsilon_b.
\end{align*}

For the second part $\dot{e}_{T_G}=\tD\exp_{\SE_2(3)}({\varepsilon_T}^{\wedge})[{\dot{\varepsilon}_{T_G}}^{\wedge}]$, one has 
\begin{align}\label{eq:tg_tg}
    e_T \frac{\text{I}-\exp(-\ad_{{\varepsilon_T}^\wedge})}{\ad_{{\varepsilon_T}^\wedge}} {\dot{\varepsilon}_{T_G}}^\wedge & = \begin{bmatrix}
        \Vector{}{0}{3\times3} & \Vector{}{g}{}-e_R\Vector{}{g}{} & e_v \\
        \Vector{}{0}{1\times3} & 0 & 0 \\
        \Vector{}{0}{1\times3} & 0 & 0
    \end{bmatrix}.
\end{align}
Multiply both sides of \equref{tg_tg} by ${e_T}^{-1}$ and then apply $\Ad_{e_T}$ to both sides:
\begin{align*}
\Ad_{e_T} \frac{\text{I}-\exp(-\ad_{\varepsilon_T})}{\ad_{\varepsilon_T}} \dot{\varepsilon}_{T_G} & = \Ad_{e_T} {e_T}^{-1}\begin{bmatrix}
    \Vector{}{0}{3\times3} & \Vector{}{g}{}-e_R\Vector{}{g}{} & e_v \\
    \Vector{}{0}{1\times3} & 0 & 0 \\
    \Vector{}{0}{1\times3} & 0 & 0    
\end{bmatrix}\\
&=\begin{bmatrix}
    \Vector{}{0}{3\times3} & \Vector{}{g}{}-e_R\Vector{}{g}{} & e_v \\
    \Vector{}{0}{1\times3} & 0 & 0 \\
    \Vector{}{0}{1\times3} & 0 & 0    
\end{bmatrix}.
\end{align*}
Use the result from \equref{jl}:
\begin{align}\label{eq:tg:temp2}
    J_l(\varepsilon_T)\dot{\varepsilon}_{T_G} = \begin{bmatrix}
        \Vector{}{0}{3\times3} & (\text{I}-e_R)\Vector{}{g}{} & J_l(\varepsilon_R)\varepsilon_v \\
        \Vector{}{0}{1\times3} & 0 & 0 \\
        \Vector{}{0}{1\times3} & 0 & 0    
    \end{bmatrix}.
\end{align}
Note that:
\begin{align*}
\text{I}-e_R &= \text{I}-\exp({\varepsilon_R}^\wedge)\\
    &= \text{I} - \sum_{k=0}^{\infty}\frac{1}{k!}{{\varepsilon_R}^\wedge}^k\\
    &= -(\sum_{k=0}^{\infty}\frac{1}{(k+1)!}{{\varepsilon_R}^\wedge}^k){\varepsilon_R}^\wedge\\
    &= -J_l(\varepsilon_R){\varepsilon_R}^\wedge.
\end{align*}
One can then simplify \equref{tg:temp2} to 
\begin{align*}
    \dot{\varepsilon}_{T_G} = \begin{bmatrix}
        \Vector{}{0}{3\times3} & \Vector{}{g}{}^\wedge\varepsilon_R & \varepsilon_v \\
        \Vector{}{0}{1\times3} & 0 & 0 \\
        \Vector{}{0}{1\times3} & 0 & 0    
    \end{bmatrix}.
\end{align*}
Combining the results for $\dot{\varepsilon}_{T_W}$ and $\dot{\varepsilon}_{T_G}$, one has 
\begin{align*}
    \dot{\varepsilon}_{T} = (\varepsilon_{b_\omega}, \;\varepsilon_{b_a}+\Vector{}{g}{}^\wedge\varepsilon_R, \;\varepsilon_{b_\nu}+\varepsilon_v)^\wedge.
\end{align*}

\paragraph*{Bias states}
The linearization of the bias error are derived from the formula given by 
\begin{align*}
    &\dot{\varepsilon} = \mathbf{A}_{t}^{0}\varepsilon + \mathcal{O}(\varepsilon^2) ,\\
    &\mathbf{A}_{t}^{0} = \Fr{e}{\mathring{\xi}}\vartheta\left(e\right)\Fr{E}{\text{I}}\phi_{\mathring{\xi}}\left(E\right)\Fr{e}{\mathring{\xi}}\Lambda\left(e, {u^\circ}\right)\Fr{\varepsilon}{\mathbf{0}}\vartheta^{-1}\left(\varepsilon\right).
\end{align*}
In this case, because we choose normal coordinates as the local coordinate chart, that is, $\vartheta \coloneqq  \log \circ\; \phi_{\mathring{\xi}}^{-1}$, we have
\begin{align*}
    \dot{\varepsilon} = \Fr{e}{\mathring{\xi}}\Lambda\left(e, \mathring{u}\right) \Fr{E}{\text{I}}\phi_{\mathring{\xi}}(E)[\varepsilon] + \mathcal{O}(\varepsilon^2).
\end{align*}
Evaluating $\tD\phi_{\mathring{\xi}}$ at $\text{I}$ with direction $[\varepsilon_T, \varepsilon_b]$ yields
\begin{align*}
    \tD\phi_{\mathring{\xi}}(\text{I})[\varepsilon_T,\varepsilon_b] = ({\varepsilon_T}^\wedge, -{\varepsilon_b}^\wedge).
\end{align*}
Evaluating $\tD\Lambda_{\mathring{u}}$ at $\mathring{\xi}$ with direction $[{\varepsilon_T}^\wedge, -{\varepsilon_b}^\wedge]$ yields
\begin{align*}
    \tD\Lambda_{\mathring{u}}(\mathring{\xi})[{\varepsilon_T}^\wedge, -{\varepsilon_b}^\wedge] &= 
    (\sim, \adsym{-{\varepsilon_b}^\wedge}{\Lambda_1(\mathring{\xi}, \mathring{u})})\\
    &=(\sim, \adMsym{\left(\mathring{\Vector{}{w}{}}^{\wedge} + \mathbf{G}\right)}\varepsilon_b).
\end{align*}
Hence the linearization of bias error is given by
\begin{align*}
    \dot{\varepsilon}_b = \adMsym{\left(\mathring{\Vector{}{w}{}}^{\wedge} + \mathbf{G}\right)}\varepsilon_b + \mathcal{O}(\varepsilon^2).
\end{align*}

\subsubsection{Filter design}
The linearized error state matrix ${\mathbf{A}_{t}^{0} \st \dotVector{}{\varepsilon}{} \simeq \mathbf{A}_{t}^{0}\Vector{}{\varepsilon}{}}$ is defined according to

\begin{equation}
    \mathbf{A}_{t}^{0} = \begin{bNiceArray}{w{c}{0,75cm}:w{c}{1,75cm}}[margin]
        \prescript{}{2}{\mathbf{A}} & \eye_9\Bstrut\\
        \hdottedline
        \Vector{}{0}{9 \times 9} & \adMsym{\left(\mathring{\Vector{}{w}{}}^{\wedge} + \mathbf{G}\right)}\Tstrut
    \end{bNiceArray} \in \R^{18 \times 18}, \label{eq:At0_tg}
\end{equation}
with $\prescript{}{2}{\mathbf{A}}$ in \equref{2A}.

Position measurements formulated according to \equref{confout_vec} are equivariant, yielding the following output matrix
\begin{equation}
    \mathbf{C}^{\star} = \begin{bmatrix}
        \frac{1}{2}\left(y + \hatVector{}{p}{}\right)^{\wedge} & \mathbf{0}_{3 \times 3} & -\eye_3 & \mathbf{0}_{3 \times 9}
        \end{bmatrix} \in \R^{3 \times 18}. \label{eq:Cstar_tg}
\end{equation}
Moreover, an additional constraint can be imposed on the virtual bias ${\Vector{}{b}{\nu}}$; that is, an additional measurement in the form of ${h\left(\xi\right) = \Vector{}{b}{\nu} = \Vector{}{0}{} \in \R^{3}}$ can be considered, leading to the following output matrix
\begin{equation}
    \mathbf{C}^{0} = \begin{bmatrix}
        \mathbf{0}_{3 \times 3} & \mathbf{0}_{3 \times 3} & \mathbf{0}_{3 \times 3} & -\hatRot{}{}^\top & \mathbf{0}_{3 \times 3} & \hatRot{}{}^\top \hatVector{}{p}{}^{\wedge}
        \end{bmatrix} \in \R^{3 \times 18}. \label{eq:C0_bnu}
\end{equation}

Note that for a practical implementation of the presented \ac{eqf} the virtual inputs ${\Vector{}{\nu}{}}$ is set to zero.

It is worth noticing that the \ac{eqf} built on the $\grpD$ symmetry group is the only filter with exact linearization of the navigation error dynamics.

\subsection{DP-\ac{eqf}}
\subsubsection{Overview}
The state space is defined as ${\calM \coloneqq \mathcal{HG}(3)\times\R^3\times\R^6}$ with ${\xi \coloneqq (\Pose{}{},\Vector{}{p}{},\Vector{}{b}{})\in\calM}$. 
One has ${\Pose{}{} = (\Rot{}{},\Vector{}{v}{})\in\mathcal{HG}(3)}$ and ${\Vector{}{b}{}=(\Vector{}{b}{\bm{\omega}},\Vector{}{b}{a})\in\R^6}$.
Choose the origin to be ${\mathring{\xi} =  \left(\eye_4,\Vector{}{0}{6\times1},\Vector{}{0}{3\times1}\right)\in\calM}$.
The velocity input is given by ${u \coloneqq \left(\Vector{}{\omega}{}, \Vector{}{a}{},\Vector{}{\tau}{\bm{\omega}},\Vector{}{\tau}{a}, \Vector{}{\nu}{}\right)}$.

The symmetry group of DP-\ac{eqf} is given by ${\grpG_{\mathbf{DP}}:\mathbf{HG}(3)\ltimes \gothhg(3) \times \R^3}$. Define the filter state ${\hat{X} = (\hat{B},\hat{\beta},\hat{c})\in\grpG_{\mathbf{DP}}}$ with ${\hat{B} = (\hat{A},\hat{a})\in\mathbf{HG}(3)}$ and ${\hat{\beta} = (\hat{\beta_\omega}, \hat{\beta_a})^\wedge \in \gothhg(3)}$.
The state estimate is given by 
\begin{align}
    \hat{\xi} \coloneqq \phi(\hat{X},\mathring{\xi}) = (\hat{B}, \Ad_{\hat{B}^{-1}}(-\hat{\beta}),\hat{c}) = (\hatPose{}{}, \hatVector{}{b}{}, \hatVector{}{p}{}). 
\end{align}
The state error is defined as 
\begin{align}\label{eq:dp_err}
e \coloneqq \phi(\hat{X}^{-1},\xi) &= (\Pose{}{}\hat{B}^{-1},\AdMsym{\hat{B}}\Vector{}{b}{} + \hat{\beta}^{\vee}, \Vector{}{p}{} - \hat{c})\\
&= (\Pose{}{}\hatPose{}{}^{-1},\AdMsym{\hatPose{}{}}(\Vector{}{b}{} - \hatVector{}{b}{}), \Vector{}{p}{} - \hatVector{}{p}{})
\end{align}

\subsubsection{Error dynamics}
\paragraph*{Navigation states}
Because of the semi-direct product structure related to the rotation and velocity states and the corresponding bias states, the derivation of the error dynamics of the $\mathcal{HG}(3)$ part is similar to the one in TG-\ac{eqf}.
In this case, one has 
\begin{align*}
    &\dot{\varepsilon}_R = \varepsilon_{b_\omega},\\
    &\dot{\varepsilon}_v = \varepsilon_{b_a}+\Vector{}{g}{}^\wedge\varepsilon_R.
\end{align*}

For the position error, one has 
\begin{align*}
    \dot{\varepsilon}_p &= \dot{e}_p = \dot{\Vector{}{p}{}} - \dot{\hatVector{}{p}{}} = \Rot{}{}\Vector{}{\nu}{} + \Vector{}{v}{} - \hatRot{}{}\Vector{}{\nu}{} - \hatVector{}{v}{}\\
    &=e_v+e_R\hatVector{}{v}{}-\hatVector{}{v}{}+e_R\hatRot{}{}\Vector{}{\nu}{}-\hatRot{}{}\Vector{}{\nu}{}\\
    &=J_l(\varepsilon_R)\varepsilon_v+(\text{I}+\varepsilon_R^\wedge + O({\varepsilon_R}^2))\left(\hatRot{}{}\Vector{}{\nu}{}+\hatVector{}{v}{}\right)-\left(\hatRot{}{}\Vector{}{\nu}{}+\hatVector{}{v}{}\right)\\
    &=J_l(\varepsilon_R)\varepsilon_v+(\text{I}+\varepsilon_R^\wedge + O({\varepsilon_R}^2))\mathring{\Vector{}{\nu}{}}-\mathring{\Vector{}{\nu}{}}\\
    &=\varepsilon_v - \mathring{\Vector{}{\nu}{}}^\wedge\varepsilon_R + \mathcal{O}(\varepsilon^2).
\end{align*}

\paragraph*{Bias states}
The derivation of bias error dynamics is the same as TG-\ac{eqf}:
\begin{align*}
    \dot{\varepsilon}_b = \adMsym{\left(\mathring{\Vector{}{w}{}}^{\wedge} + \mathbf{G}\right)}\varepsilon_b + \mathcal{O}(\varepsilon^2).
\end{align*}

\subsubsection{Filter design}
The linearized error state matrix ${\mathbf{A}_{t}^{0} \st \dotVector{}{\varepsilon}{} \simeq \mathbf{A}_{t}^{0}\Vector{}{\varepsilon}{}}$ is defined according to
\begin{equation}
    \mathbf{A}_{t}^{0} = \begin{bmatrix}
        \prescript{}{3}{\mathbf{A}} & \eye_6 & \mathbf{0}_{6 \times 3}\\
        \mathbf{0}_{6 \times 6} & \adMsym{\left(\mathring{\Vector{}{w}{}}^{\wedge} + \mathbf{G}\right)^{\vee}} & \mathbf{0}_{6 \times 3}\\
        \prescript{}{4}{\mathbf{A}} & \mathbf{0}_{3 \times 6} & \mathbf{0}_{3 \times 3}\\
        \end{bmatrix} \in \R^{15 \times 15}, \label{eq:At0_SE3_R3}
\end{equation}
where
\begin{align*}
    &\prescript{}{3}{\mathbf{A}} = \begin{bmatrix}
    \Vector{}{0}{3 \times 3} & \Vector{}{0}{3 \times 3}\\
    \Vector{}{g}{}^{\wedge} & \Vector{}{0}{3 \times 3}
    \end{bmatrix} \in \R^{6 \times 6}\\
    &\prescript{}{4}{\mathbf{A}} = \begin{bmatrix}
    -\mathring{\Vector{}{\nu}{}}^{\wedge} & \eye_3
    \end{bmatrix} \in \R^{3 \times 6}.
\end{align*}

Position measurements are linear; therefore, the output matrix is written
\begin{equation}
    \mathbf{C}^{0} = \begin{bmatrix}
        \mathbf{0}_{3 \times 3} & \mathbf{0}_{3 \times 3} & \mathbf{0}_{3 \times 6} & \eye_3\\
        \end{bmatrix} \in \R^{3 \times 15}. \label{eq:C0_SE3_R3}
\end{equation}

Similar to the previous filter, for a practical implementation of the presented \ac{eqf}, the virtual input ${\Vector{}{\nu}{}}$ is set to zero.

\subsection{SD-\ac{eqf}}
\subsubsection{Overview}
The state space is defined as ${\calM \coloneqq \mathcal{SE}_2(3)\times\R^6}$ with ${\xi \coloneqq (\Pose{}{},\Vector{}{b}{})\in\calM}$. 
One has ${\Pose{}{} = (\Rot{}{},\Vector{}{v}{},\Vector{}{p}{})\in\mathcal{SE}_2(3)}$ and ${\Vector{}{b}{}=(\Vector{}{b}{\bm{\omega}},\Vector{}{b}{a}) \in \R^6}$.
Choose the origin to be ${\mathring{\xi} = \left(\eye_5,\Vector{}{0}{6\times1}\right)\in\calM}$.
The velocity input is given by ${u \coloneqq \left(\Vector{}{\omega}{}, \Vector{}{a}{}, \Vector{}{\tau}{\bm{\omega}}, \Vector{}{\tau}{a}\right)}$.

The symmetry group of SD-\ac{eqf} is given by ${\grpG_{\mathbf{SD}}:\SE_2(3)\ltimes \se(3)}$. Define the filter state ${\hat{X} = (\hat{C},\hat{\gamma})\in\grpG_{\mathbf{SD}}}$ with ${\hat{C} = (\hat{A},\hat{a},\hat{b})\in\SE_2(3)}$ and ${\hat{\gamma} = (\hat{\gamma_\omega}, \hat{\gamma_a})^\wedge\in\se(3)}$. The $\SE_2(3)$ component in $\hat{X}$ can also be expressed in $\hat{C} = (\hat{B},\hat{b})$ where $\hat{B} = (\hat{A},\hat{a})\in\mathbf{HG}(3)$.
The state estimate is given by 
\begin{align}
    \hat{\xi} \coloneqq \phi(\hat{X},\mathring{\xi}) = (\hat{C}, \Ad_{\hat{B}^{-1}}(-\hat{\gamma}^\vee)) = (\hatPose{}{}, \hatVector{}{b}{}). 
\end{align}
The state error is defined as 
\begin{align}\label{eq:sd_err}
e \coloneqq \phi(\hat{X}^{-1},\xi) &= (\Pose{}{}\hat{C}^{-1},\AdMsym{\hat{B}}(\Vector{}{b}{}+\Adsym{\hat{B}^{-1}}{\hat{\gamma}}^\vee))\\
&= (\Pose{}{}\hat{C}^{-1},\AdMsym{\hat{B}}\Vector{}{b}{} + \hat{\gamma}^{\vee}).
\end{align}

\subsubsection{Error dynamics}
\paragraph*{Navigation states}
Because of the semi-direct product structure related to the rotation and velocity states and the corresponding bias states, the derivation of the error dynamics of the $\mathbf{HG}(3)$ part is similar to the one in TG-\ac{eqf}.
In this case, one has 
\begin{align*}
    &\dot{\varepsilon}_R = \varepsilon_{b_\omega},\\
    &\dot{\varepsilon}_v = \varepsilon_{b_a}+\Vector{}{g}{}^\wedge\varepsilon_R.
\end{align*}

For the position error $\dot{e}_p=\dot{\varepsilon}_p+\mathcal{O}(\varepsilon^2)$, one has 
\begin{align*}
    \dot{e}_p &= \ddt (-\Rot{}{}\hatRot{}{}^\top\hatVector{}{p}{}+\Vector{}{p}{})\\
        &= -\dot{e}_R\hatVector{}{p}{} - e_R\dot{\hatVector{}{p}{}}+\dot{\Vector{}{p}{}}\\
        &= e_Re_{b_\omega}^\wedge \hatVector{}{p}{} - e_R\hatVector{}{v}{} + \Vector{}{v}{}\\
    \dot{\varepsilon}_p &= ((\text{I}+\varepsilon_R^\wedge + O({\varepsilon}^2)))(\varepsilon_{b_\omega}^\wedge+\mathcal{O}(\varepsilon^2))\hatVector{}{p}{}+(\varepsilon_v+\mathcal{O}(\varepsilon^2))\\
    &=\varepsilon_v+\hatVector{}{p}{}^\wedge\varepsilon_{b_\omega}+\mathcal{O}(\varepsilon^2).
\end{align*}

\paragraph*{Bias states}
The derivation of bias error dynamics is the same as TG-\ac{eqf}, and yields:
\begin{align*}
    \dot{\varepsilon}_b = \adMsym{\left(\AdMsym{\hat{B}}\Vector{}{w}{} + \hat{\gamma}^{\vee} + \mathbf{G}^{\vee}\right)}\varepsilon_b + \mathcal{O}(\varepsilon^2).
\end{align*}
Note that the following relation holds
\begin{equation*}
    \AdMsym{\hat{B}}\Vector{}{w}{} + \hat{\gamma}^{\vee} = \Pi\left(\mathring{\Vector{}{w}{}}^{\wedge}\right)^{\vee},
\end{equation*}

\subsubsection{Filter dseign}
The linearized error state matrix ${\mathbf{A}_{t}^{0} \st \dotVector{}{\varepsilon}{} \simeq \mathbf{A}_{t}^{0}\Vector{}{\varepsilon}{}}$ is defined according to
\begin{equation}
    \mathbf{A}_{t}^{0} = \begin{bNiceArray}{cw{c}{0.45cm}c:w{c}{1.15cm}w{c}{1.35cm}}[margin]
        \Block{3-3}{\prescript{}{2}{\mathbf{A}}} & & & \Block{2-2}{\eye_6} & \\
        & & & & \\
        & & & \hatVector{}{p}{}^{\wedge} & \mathbf{0}_{3 \times 3} \Bstrut\\
        \hdottedline
        \Block{1-3}{\mathbf{0}_{6 \times 9}} & & & \Block{1-2}{ \adMsym{\left(\AdMsym{\hat{B}}\Vector{}{w}{} + \hat{\gamma}^{\vee} + \mathbf{G}^{\vee}\right)}} & \Tstrut\\
    \end{bNiceArray} \in \R^{15 \times 15}, \label{eq:At0_SE23}
\end{equation}

When comparing ${\mathbf{A}_{t}^{0}}$ in \equref{At0_SE23} with the one in \equref{At0_tg}, it is trivial to see the only difference between the two matrices is in the row of ${\mathbf{A}_{t}^{0}}$ relative to the position error. This is where the major difference between filters employing the symmetries $\grpD$, and $\grpB$ is found.

Position measurements formulated according to \equref{confout_vec} are equivariant, yielding the following output matrix\begin{equation}
    \mathbf{C}^{\star} = \begin{bmatrix}
        \frac{1}{2}\left(y + \hatVector{}{p}{}\right)^{\wedge} & \mathbf{0}_{3 \times 3} & -\eye_3 & \mathbf{0}_{3 \times 6}\\
        \end{bmatrix} \in \R^{3 \times 15}. \label{eq:Cstar_SE3}
\end{equation}

\subsection{\ac{tfgiekf}}
\subsubsection{Overview}
The state space is defined as ${\calM \coloneqq \mathcal{SE}_2(3)\times\R^6}$ with ${\xi \coloneqq (\Pose{}{},\Vector{}{b}{})\in\calM}$. 
One has ${\Pose{}{} = (\Rot{}{},\Vector{}{v}{},\Vector{}{p}{})\in\mathcal{SE}_2(3)}$ and ${\Vector{}{b}{}=(\Vector{}{b}{\bm{\omega}},\Vector{}{b}{a}) \in \R^6}$.
Choose the origin to be ${\mathring{\xi} = \left(\eye_5,\Vector{}{0}{6\times1}\right)\in\calM}$.
The velocity input is given by ${u \coloneqq \left(\Vector{}{\omega}{}, \Vector{}{a}{},\Vector{}{\tau}{\bm{\omega}},\Vector{}{\tau}{a}\right)}$.

The symmetry group of TFG-IEKF is given by ${\grpG_{\mathbf{TF}}:\SO(3)\ltimes(\R^6\oplus\R^6)}$. Define the filter state ${\hat{X} = (\hat{C},\hat{\gamma})\in\grpG_{\mathbf{TF}}}$ with ${\hat{C} = (\hat{A},(\hat{a},\hat{b}))\in\SE_2(3)=\SO(3)\ltimes\R^6}$ and ${\hat{\gamma} = (\hat{\gamma_\omega}, \hat{\gamma_a})\in\R^6}$.
The state estimate is given by 
\begin{align}
    \hat{\xi} \coloneqq \phi(\hat{X},\mathring{\xi}) = (\hat{C}, \hat{A}^{-1}*(-\hat{\gamma})) = (\hatPose{}{}, \hatVector{}{b}{}). 
\end{align}
The state error is defined as 
\begin{align}\label{eq:tf_err}
e \coloneqq \phi(\hat{X}^{-1},\xi) &= (\Pose{}{}\hat{C}^{-1},\hat{A}*(\Vector{}{b}{}+\hat{A}^{-1}*\hat{\gamma})) \\
&= (\Pose{}{}\hatPose{}{}^{-1}, \hatRot{}{}*(\Vector{}{b}{} - \hatVector{}{b}{})).
\end{align}

\subsubsection{Error dynamics}
\paragraph*{Navigation states}
Because of the semi-direct product structure related to the rotation state, the error dynamics of the rotation part is similar to the TG-\ac{eqf}, but with different signs due to the use of log-coordinates $\varepsilon = \log_\grpG$:
\begin{align*}
    &\dot{\varepsilon}_R = -\varepsilon_{b_\omega}.
\end{align*}

For the velocity error $e_v = -\Rot{}{}\hatRot{}{}^\top\hatVector{}{v}{}+\Vector{}{v}{}$, one has 
\begin{align*}
    \dot{e}_v
        &=-\dot{e}_R\hatVector{}{v}{} - e_R\dot{\hatVector{}{v}{}}+\dot{v}\\
        &=e_R(e_{b_\omega})^\wedge\hatVector{}{v}{}-e_R\hatRot{}{}(\Vector{}{a}{}-\hatVector{}{b}{a}) -e_R \Vector{}{g}{} + \Rot{}{}(\Vector{}{a}{}-\Vector{}{b}{a})+\Vector{}{g}{}\\
        &=e_R(e_{b_\omega})^\wedge\hatVector{}{v}{}-e_R\hatRot{}{}(\Vector{}{a}{}-\hatVector{}{b}{a}) -(e_R-\text{I}) \Vector{}{g}{} \\
        &\quad+ e_R\hatRot{}{}(\Vector{}{a}{}-\hatRot{}{}^\top(e_{b_a}-\gamma_{b_a}))\\
        &=e_Re_{b_\omega}^\wedge\hatVector{}{v}{} - e_Re_{b_a} - (e_R-\text{I})\Vector{}{g}{};\\
    \dot{\varepsilon}_v &= -\hatVector{}{v}{}^\wedge\varepsilon_{b_\omega}+\Vector{}{g}{}^\wedge\varepsilon_R - \varepsilon_{b_a}+\mathcal{O}(\varepsilon^2).
\end{align*}

The derivation for position error $e_p = -\Rot{}{}\hatRot{}{}^\top\hatVector{}{p}{}+\Vector{}{p}{}$ is similar to the SD-\ac{eqf}, and it is given by
\begin{align*}
    \dot{\varepsilon}_p = \varepsilon_v - \hatVector{}{p}{}^\wedge\varepsilon_{b_\omega}+\mathcal{O}(\varepsilon^2).
\end{align*}

\paragraph*{Bias states}
The error in bias state ${b_\omega}$ is given by $e_{b_\omega} = \hatRot{}{}*(\Vector{}{b}{\bm{\omega}}-\hatVector{}{b}{\bm{\omega}})$.
The dynamics can be derived:
\begin{align*}
    \dot{e}_{b_\omega} &= \hatRot{}{}(\omega-\hatVector{}{b}{\bm{\omega}})^\wedge * (\Vector{}{b}{\bm{\omega}}-\hatVector{}{b}{\bm{\omega}})\\
    &=\hatRot{}{}(\omega - \hatVector{}{b}{\bm{\omega}})^\wedge\hatRot{}{}^\top\hatRot{}{}* (\Vector{}{b}{\bm{\omega}}-\hatVector{}{b}{\bm{\omega}})\\
    &=(\hatRot{}{}(\omega-\hatVector{}{b}{\bm{\omega}}))^\wedge e_{b_\omega}.
\end{align*}
In local coordinates, the linearization is given by 
\begin{align*}
    \dot{\varepsilon}_{b_\omega}=(\hatRot{}{}(\omega-\hatVector{}{b}{\bm{\omega}}))^\wedge \varepsilon_{b_\omega} + \mathcal{O}(\varepsilon^2).
\end{align*}
The error in $b_a$ follows the same derivation, which is given by 
\begin{align*}
    \dot{\varepsilon}_{b_a}=(\hatRot{}{}(\omega-\hatVector{}{b}{\bm{\omega}}))^\wedge \varepsilon_{b_a} + \mathcal{O}(\varepsilon^2).
\end{align*}

\subsubsection{Filter design}
The linearized error state matrix ${\mathbf{A}_{t}^{0} \st \dotVector{}{\varepsilon}{} \simeq \mathbf{A}_{t}^{0}\Vector{}{\varepsilon}{}}$ is defined according to
\begin{equation}
    \mathbf{A}_{t}^{0} = \begin{bNiceArray}{cw{c}{0.45cm}c:w{c}{2,75cm}w{c}{2.15cm}}[margin]
        \Block{3-3}{\prescript{}{4}{\mathbf{A}}} & & & -\eye_3 & \mathbf{0}_{3 \times 3}\\
        & & & -\hatVector{}{v}{}^{\wedge} & -\eye_{3}\\
        & & & -\hatVector{}{p}{}^{\wedge} & \mathbf{0}_{3 \times 3}\Bstrut\\
        \hdottedline
        \Block{1-3}{\mathbf{0}_{3 \times 9}} & & & \left(\hatRot{}{}\left(\Vector{}{\omega}{} - \hatVector{}{b}{\omega}\right)\right)^{\wedge} & \mathbf{0}_{3 \times 3}\Tstrut\\
        \Block{1-3}{\mathbf{0}_{3 \times 9}} & & & \mathbf{0}_{3 \times 3} & \left(\hatRot{}{}\left(\Vector{}{\omega}{} - \hatVector{}{b}{\omega}\right)\right)^{\wedge}
    \end{bNiceArray} \in \R^{15 \times 15}. \label{eq:At0_TFG}
\end{equation}

Position measurements formulated according to \equref{confout_vec} are equivariant, yielding the following output matrix
\begin{equation}
    \mathbf{C}^{\star} = \begin{bmatrix}
        \frac{1}{2}\left(y + \hatVector{}{p}{}\right)^{\wedge} & \mathbf{0}_{3 \times 3} & -\eye_3 & \mathbf{0}_{3 \times 6}\\
        \end{bmatrix} \in \R^{3 \times 15}. \label{eq:Cstar_tfgiekf}
\end{equation}